\documentclass{article}

\usepackage[preprint]{neurips_2026}

\usepackage[utf8]{inputenc} 
\usepackage[T1]{fontenc}    
\usepackage{hyperref}       
\usepackage{url}            
\usepackage{booktabs}       
\usepackage{amsfonts}       
\usepackage{nicefrac}       
\usepackage{microtype}      
\usepackage{xcolor}         
\usepackage{booktabs}
\usepackage{multirow}
\usepackage{graphicx}
\usepackage{amsmath} 
\usepackage[most]{tcolorbox}
\usepackage{enumitem}
\usepackage{xcolor}
\tcbuselibrary{listings,breakable}
\usepackage[most]{tcolorbox}
\tcbuselibrary{listings,breakable}
\title{RaMem: Contextual Reinstatement for \\ Long-term Agentic Memory}

\author{%
  \textbf{Wei Yang, Bryce Kan, Shixuan Li, Li Li, Yuehan Qin,} \\
  \textbf{Jiate Li, Paul Bogdan, Jesse Thomason} \\
  University of Southern California \\
  \texttt{\{wyang930,brycekan,sli97750,pbogdan,jessetho\}@usc.edu}
}

\begin{document}

\maketitle

\begin{abstract}
Long-term memory has become increasingly important for LLM agents that operate across extended interactions and evolving task contexts. Recent memory systems have made past experiences more persistent, compact, and retrievable, but retrieval alone does not ensure that a memory provides valid evidence for the current query. When experiences are compressed into reusable fragments, memories from different situations may appear equally relevant if they involve recurring entities or user states. We refer to this failure as \textbf{context collapse}: memories lose the surrounding context needed to judge whether they provide valid evidence for the current query. To address this problem, we propose \textbf{Contextual Reinstatement for Agentic Memory (RaMem)}, a framework that turns retrieved memory fragments into contextually verifiable evidence. RaMem operates through four coordinated stages: (\emph{i}) \textbf{evidence anchoring} grounds each memory in its original episodic conditions, especially event time, mention time, session span, and participants; (\emph{ii}) \textbf{recall condition induction} derives the evidence conditions implied by the query; (\emph{iii}) \textbf{validity-aware retrieval} uses these conditions to prioritize context-compatible memories while retaining content-relevant candidates as fallback evidence; and (\emph{iv}) \textbf{context-preserved synthesis} keeps the selected memories' structured context available to the generator. Experiments on long-term memory benchmarks show that RaMem consistently improves performance over strong memory baselines, with average F1 gains of more than 10\% across several backbones. Code is available at \url{https://github.com/weiyang930/RaMem-Release.git}.
\end{abstract}

\section{Introduction}

Long-term memory has become a central component of agentic systems, allowing agents to carry useful experience across interactions and reason beyond a single context window. Existing work has developed memory along several complementary directions. Some systems maintain persistent natural-language records, virtual contexts, or user-level facts across sessions~\citep{park2023generativeagentsinteractivesimulacra,packer2024memgptllmsoperatingsystems,zhong2024memorybank,chhikara2025mem0}. Others organize memory with temporal graphs, linked notes, symbolic metadata, or multi-view indexes to support consolidation and retrieval~\citep{rasmussen2025zep,liu2026simplemem}. Recent agentic memory systems further study how memory should be actively managed during task execution, including when to distill experience, reuse procedures, retrieve past information, or update stored knowledge~\citep{ma2025deserves,cao2025remember,yu2026agentic,du2025memr}. Together, these efforts help agents maintain usable memory beyond transient context storage. At a high level, they largely frame memory as an \emph{availability problem}: what information should be stored, and how it should be retrieved when needed.

\begin{figure*}[t]
\centering
\includegraphics[width=\textwidth]{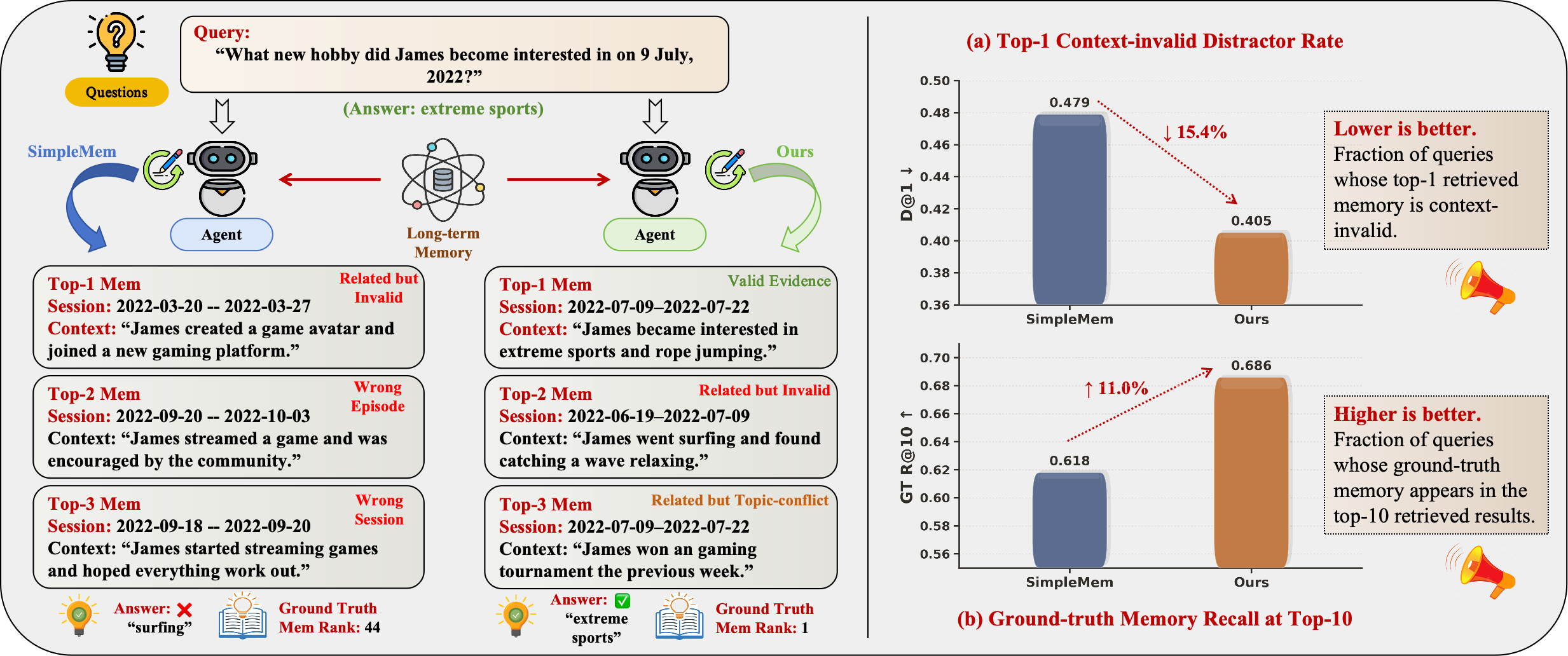}
\caption{Context collapse in long-term agent memory. \textbf{Left}: the example shows the core failure mode of context collapse. SimpleMem retrieves related but invalid memories from wrong sessions. \textbf{Right}: aggregate results on the context-confusable subset show that this failure is systematic. SimpleMem often promotes context-invalid memories to rank 1 and misses the ground-truth memory.}
\label{fig:intro_context_collapse}
\end{figure*}

Availability, however, does not guarantee that a retrieved memory can be reliably used as evidence. Most memory systems implicitly rely on what we call the \emph{decontextualized evidence assumption}: once an experience is compressed into a memory fragment and retrieved through a relevance signal, it can serve as evidence for future queries~\citep{chhikara2025mem0,liu2026simplemem}.
Long-term experience is not a collection of context-free fragments~\citep{tulving2002episodic}. A memory may depend on when it was observed, who was involved, what state the user or task was in, where it occurred in an event, and how it related to nearby interactions. When this context is weakened during compression, storage, or retrieval, multiple fragments may appear relevant to the same query even though only one is valid for the situation being asked about\citep{liu2025echolargelanguagemodel}. We call this failure mode \textbf{context collapse}: the surrounding context of an experience is flattened, causing related but context-invalid memories to compete as evidence. Context collapse is not simply a retrieval miss, but a form of \textbf{evidence misidentification}, where the agent retrieves information that is related to the query but belongs to the wrong context. Figure~\ref{fig:intro_context_collapse} illustrates this gap between retrieved relevance and evidential validity.

We argue that a key missing property of long-term agent memory is \textbf{contextual verifiability}. A memory should not only be stored and retrieved; it should retain enough context for the agent to judge whether it can support the current query. For example, answering a query about a revised plan requires identifying which version of the plan is under discussion, while inferring a user preference requires determining whether the memory reflects a current preference, an earlier state, or a temporary exception. This view is consistent with the idea that memory retrieval depends on both content and context. Episodic memory and encoding specificity emphasize that retrieval is shaped by the match between a cue and the context in which the experience was encoded~\citep{tulving2002episodic,tulving1973encoding}; work on temporal context and event segmentation further highlights the role of contextual states and event boundaries in organizing memory~\citep{howard2002distributed,kurby2008segmentation}. We draw on these ideas as a functional principle for agent memory: a retrieved memory should be treated as evidence only after the relevant context has been reinstated.

To make retrieved memories contextually verifiable, we propose \textbf{Contextual Reinstatement for Agentic Memory (RaMem)}, a framework that turns retrieved fragments into situated evidence by recovering the conditions under which each memory is applicable. RaMem implements this idea in four stages. First, \textbf{Episodic Memory Anchoring} attaches each memory to the key conditions of its original experience, including event time, mention time, session span, and participants. Second, \textbf{Recall Condition Induction} decomposes the query into an information need and a contextual recall frame that specifies what valid evidence should satisfy. Third, \textbf{Validity-Aware Retrieval} prioritizes memories whose episodic coordinates match the recall frame while retaining content-relevant candidates as fallback evidence. Finally, \textbf{Context-Preserved Synthesis} passes the selected memories to the generator with their structured context intact. Because contextual cues can be incomplete or noisy, RaMem applies reinstatement selectively and activates context-aware retrieval only when the relevant cues can be grounded reliably.


The main contributions of this work are as follows:
\begin{itemize}
    \item We identify \textbf{context collapse} as a central failure mode of long-term agent memory: memories can appear relevant but be misused as evidence when the agent cannot verify the context in which they were formed. We formulate this challenge as \textbf{contextual evidence identification}.

    \item We propose \textbf{Contextual Reinstatement for Agentic Memory (RaMem)}, a framework that turns retrieved fragments into situated evidence. RaMem makes memories contextually verifiable by anchoring them to episodic conditions, inducing recall conditions, retrieving validity-aware evidence, and preserving structured context during answer synthesis.

    \item Experiments on long-term memory benchmarks show that RaMem consistently improves answer quality, ground-truth memory retrieval, and context-budget efficiency over strong memory baselines.
\end{itemize}

\section{Related Work}
\label{sec:related_work}

Long-term memory has become an important mechanism for extending LLM agents beyond a single interaction. Early systems maintain external memory streams, virtual context, or persistent user facts, enabling agents to carry information across sessions~\citep{park2023generativeagentsinteractivesimulacra,shinn2023reflexion,packer2024memgptllmsoperatingsystems,zhong2024memorybank}. Later work makes these memories more structured by using temporal knowledge graphs, note-like organizations, graph-augmented facts, lightweight memory modules, and multi-view indexes~\citep{rasmussen2025zep,xu2025mem,chhikara2025mem0,fang2025lightmem,liu2026simplemem}. Recent methods further study how agents should distill experience, reuse procedural knowledge, learn memory operations, or control retrieval and reflection during task execution~\citep{nan2025nemori,yu2026agentic,yu2025memagent,cao2025remember,du2025memr}. These directions have made agent memory more persistent, compact, and easier to retrieve. In parallel, long-context modeling and retrieval-augmented generation improve how models access large information sources through context extension, compression, retrieval, graph-based evidence aggregation, and agentic retrieval policies~\citep{lewis2020retrieval,liu2024lost,jiang2023llmlingua,asai2023self,edge2024local,guo2024lightrag,yao2022react}. 
A more detailed discussion is provided in Appendix~\ref{app:related_work}.

\section{Method}
\label{sec:method}

In this section, we introduce \textbf{Contextual Reinstatement for Agentic Memory (RaMem)}, a framework that turns retrieved memory fragments into contextually verifiable evidence. Figure~\ref{fig:main_figure} provides an overall structure. The complete method details are provided in Appendix~\ref{app:implementation_details}.
%

\subsection{Episodic Memory Anchoring}
\label{subsec:episodic_coordinates}

A central source of context collapse is that memory units often become reusable fragments whose applicability conditions are weakly represented. Although such fragments may preserve useful content, they do not expose enough context to verify when the content can serve as evidence. RaMem therefore anchors each memory to the conditions under which it was observed, forming an \textbf{episodic evidence unit}. Each memory $m_i$ consists of a content field $x_i$ and an episodic context field $e_i$:
$
m_i = (x_i, e_i).
$
The content $x_i$ is a self-contained restatement of the remembered information, while $e_i$ records the coordinates needed for contextual verification:
\begin{equation}
e_i = \{\tau_i^{\mathrm{event}}, \tau_i^{\mathrm{mention}}, \tau_i^{\mathrm{session}}, p_i, \ell_i, u_i, z_i\}.
\end{equation}
Here $\tau_i^{\mathrm{event}}$ is the time of the described event when inferable, $\tau_i^{\mathrm{mention}}$ denotes an explicitly mentioned temporal cue when available, and $\tau_i^{\mathrm{session}}=[s_i^{\mathrm{start}},s_i^{\mathrm{end}}]$ is the session span in which the memory was observed. The remaining fields denote participants $p_i$, location $\ell_i$, entities $u_i$, and topic description $z_i$.

The distinction between event time and session time is central to evidence verification. An event may be discussed after it occurred, so $\tau_i^{\mathrm{event}}$ describes time of the real-world occurrence, while $\tau_i^{\mathrm{session}}$ records when the memory was observed in the conversation. Since many long-horizon queries depend on when information was mentioned rather than when the event happened, we treat the session span as a stable episodic coordinate. In practice, we process the interaction history with overlapping windows, extract memory entries with lossless restatements and structured cues, and match each memory back to the real session span of its source window. 

\begin{figure*}[t]
\centering
\includegraphics[width=\textwidth]{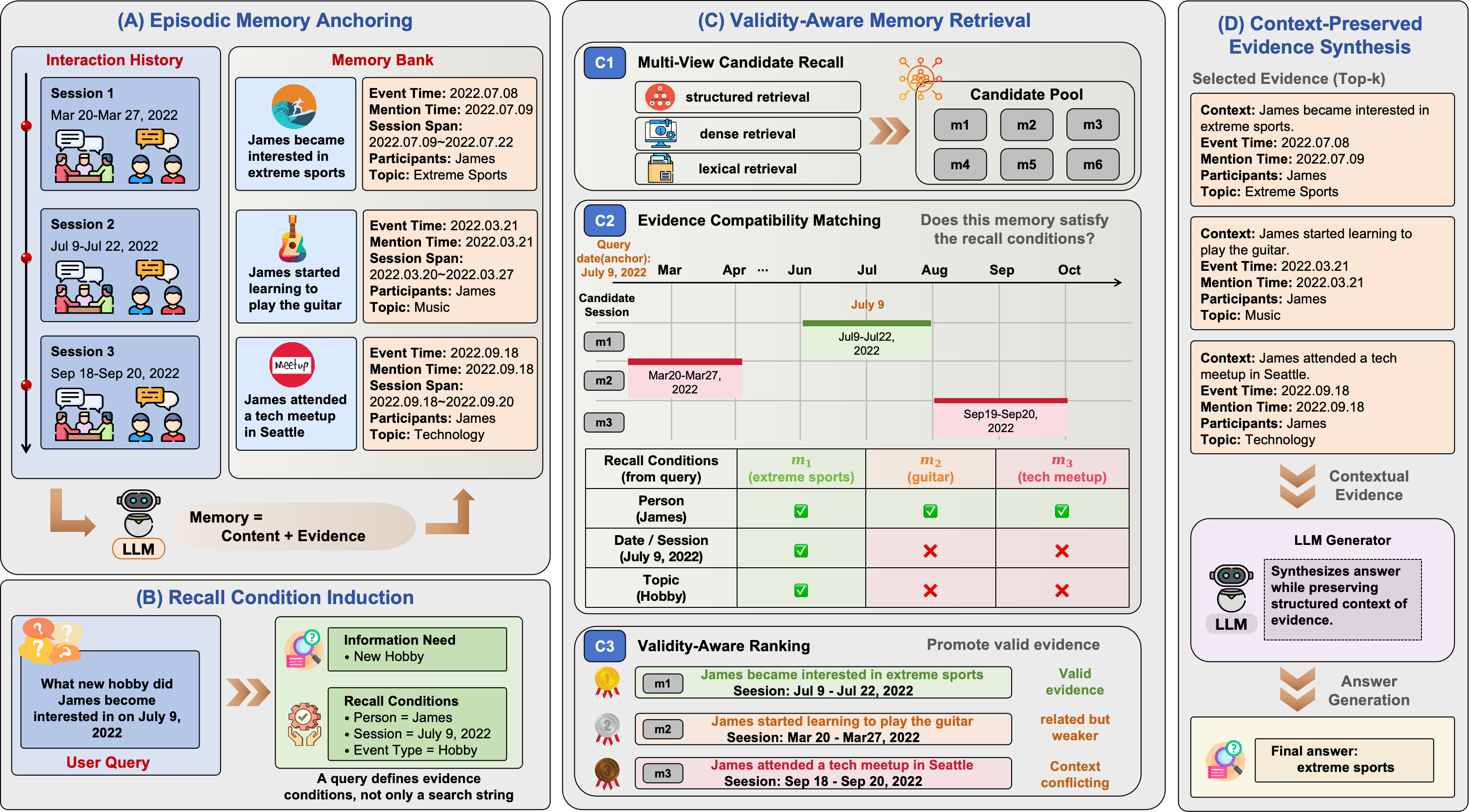}
\caption{
Overview of \textbf{RaMem}. 
RaMem converts long-term interaction history into contextually verifiable memory evidence through four stages. 
\textbf{(A)} Interaction histories are converted into memories anchored with episodic evidence conditions. 
\textbf{(B)} A query is decomposed into an information need and recall conditions. 
\textbf{(C)} RaMem retrieves candidates through multiple paths and prioritizes context-compatible evidence when grounded recall conditions are available. 
\textbf{(D)} The selected evidence is passed to the generator with structured context preserved, enabling answer synthesis from contextually verifiable memories.
}
\label{fig:main_figure}
\end{figure*}

\subsection{Recall Condition Induction}
\label{subsec:recall_frame}

A query specifies not only what information is needed, but also the conditions under which a memory can serve as valid evidence. These conditions may be either explicit or implicit: explicit anchors include named entities or dates, while implicit anchors arise from temporal cues, referenced events, or evolving user state. Therefore, we map each query $q$ into an information need $r_q$ and a contextual recall frame $c_q$:
$
q \mapsto (r_q, c_q).
$
The information need $r_q$ captures what is being requested, while $c_q$ records the recall conditions that constrain which episode can provide valid evidence:
\begin{equation}
c_q = \{\hat{\tau}_q, \hat{p}_q, \hat{\ell}_q, \hat{u}_q, \hat{z}_q\}.
\end{equation}
Here $\hat{\tau}_q$ is a grounded time range when inferable, $\hat{p}_q$ denotes people or participants, $\hat{\ell}_q$ denotes locations, $\hat{u}_q$ denotes entities, and $\hat{z}_q$ denotes topic or episode hints.

The recall frame separates the requested information from the evidence conditions that a supporting memory should satisfy. For example, in ``What did Sarah decide after the appointment in late March?'', $r_q$ concerns Sarah's decision, while $c_q$ specifies Sarah, the appointment episode, the temporal relation ``after'', and the grounded range for late March. This decomposition is necessary because a memory may answer the same type of question while belonging to a different episode. To ground relative expressions such as ``last August'' or ``the first weekend of May'', the query analyzer can use the global date range of the interaction history, making temporal cues resolvable within the memory timeline. Since induced recall conditions can still be ambiguous, RaMem does not apply them unconditionally; instead, contextual retrieval is activated only when the relevant cues can be grounded reliably, as described next.

\subsection{Validity-Aware Memory Retrieval}
\label{subsec:selective_reinstatement}

Recall conditions are useful only when they can be grounded reliably. If an inferred condition is wrong, enforcing it may exclude the correct memory and turn contextual verification into a false constraint. RaMem therefore uses \textbf{selective contextual reinstatement}: it activates validity-aware retrieval only when the recall frame contains grounded evidence conditions; otherwise, it falls back to content-based retrieval.

RaMem first retrieves a broad candidate set using complementary content-based access paths, including dense semantic retrieval over memory restatements and lexical retrieval over keywords or full-text matches. When enabled, an optional LLM-based planning step first analyzes the query and generates a small set of targeted semantic search queries while retaining the original query. When contextual retrieval is not activated, semantic and lexical results are fused with reciprocal rank fusion:
\[
\mathcal{C}_{\mathrm{content}}(q)
=
\mathrm{RRF}\big(
\mathcal{C}_{\mathrm{sem}}(q),
\mathcal{C}_{\mathrm{lex}}(q)
\big),
\]
where $\mathcal{C}_{\mathrm{sem}}(q)$ is obtained from dense retrieval over $x_i$ and $\mathcal{C}_{\mathrm{lex}}(q)$ from keyword or BM25-style retrieval. When grounded temporal conditions are available, RaMem can instead apply the same fusion inside the session-overlap filtered candidate set. This step recalls memories that may contain relevant content, but it does not by itself verify whether they satisfy the query's evidence conditions.

When the recall frame provides grounded conditions, RaMem constructs a context-compatible candidate list rather than directly treating all content-relevant memories as equally valid evidence. Let $\mathcal{C}_{\mathrm{content}}(q)$ be the content-retrieved candidates and let $\mathcal{C}_{\mathrm{ctx}}(q)$ denote memories retrieved under grounded contextual conditions. The candidate list used for evidence synthesis is
\begin{equation}
\mathcal{C}_q =
\begin{cases}
\operatorname{Dedup}_{\mathrm{first}}\big(
\mathcal{C}_{\mathrm{ctx}}(q) \oplus \mathcal{C}_{\mathrm{content}}(q)
\big), & \text{if recall conditions are grounded},\\
\mathcal{C}_{\mathrm{content}}(q), & \text{otherwise}.
\end{cases}
\end{equation}
Here $\oplus$ denotes ordered concatenation, and $\operatorname{Dedup}_{\mathrm{first}}$ keeps the first occurrence of each memory according to its identifier. Thus, grounded context gives priority to memories that are applicable under the query's evidence conditions, while content-relevant candidates are retained as fallback evidence. When the induced context is uncertain, it is not imposed as a hard constraint.

For temporal conditions, $\mathcal{V}(q)$ is instantiated with session-overlap compatibility. Given a grounded query range $\hat{\tau}_q=[t_q^{\mathrm{start}},t_q^{\mathrm{end}}]$ and a memory session span $\tau_i^{\mathrm{session}}=[s_i^{\mathrm{start}},s_i^{\mathrm{end}}]$, memory $m_i$ is temporally compatible if
\begin{equation}
s_i^{\mathrm{start}} \leq t_q^{\mathrm{end}}
\quad \text{and} \quad
s_i^{\mathrm{end}} \geq t_q^{\mathrm{start}}.
\end{equation}
A small buffer can be added around $\hat{\tau}_q$ to tolerate mismatch between event time and mention time. Within the compatible set, memories are ordered by content relevance and temporal proximity, where proximity favors sessions closer to the grounded query range. This validity-aware design directly targets context collapse: a relevance-only retriever may rank a related memory highly even when it violates the query's evidence conditions. By prioritizing context-compatible memories when reliable recall conditions are available, while retaining content-relevant candidates as fallback evidence, RaMem separates two notions that are often conflated: a memory can be related to the query while still being invalid evidence for it. After the initial retrieval results are merged, an optional reflection step can further check whether the current memories cover the required information and issue a small number of additional semantic searches for missing evidence.



\subsection{Context-Preserved Evidence Synthesis}
\label{subsec:context_stabilized_synthesis}

After validity-aware retrieval, selected memories should not be reduced back to plain text fragments. Doing so would remove the evidence conditions used to verify their applicability. RaMem therefore performs \textbf{context-preserved evidence synthesis}: it merges retrieved candidates while preserving retrieval priority, removes duplicate memories, and formats the final evidence with its episodic coordinates.

Let $\mathcal{C}_q^{\mathrm{content}}$ denote the ordered candidates obtained from content-based retrieval, and let $\mathcal{C}_q^{\mathrm{ctx}}$ denote the ordered candidates retained or promoted through contextual reinstatement. We treat both as ranked lists rather than unordered sets. RaMem first places context-compatible candidates before general content-relevant fallback evidence, and then applies stable deduplication:
\begin{equation}
\mathcal{L}_q
=
\operatorname{Dedup}_{\mathrm{first}}
\left(
\mathcal{C}_q^{\mathrm{ctx}}
\oplus
\mathcal{C}_q^{\mathrm{content}}
\right),
\end{equation}
where $\oplus$ denotes ordered concatenation, and $\operatorname{Dedup}_{\mathrm{first}}$ keeps the first occurrence of each memory according to its identifier. The final evidence list is obtained by taking the first $K$ memories from the deduplicated list:
\begin{equation}
\mathcal{C}_q
=
\operatorname{Head}_{K}(\mathcal{L}_q).
\end{equation}
When contextual reinstatement is not activated, $\mathcal{C}_q^{\mathrm{ctx}}$ is empty and the same operation reduces to content-based evidence selection. This construction preserves broad content coverage while giving priority to memories whose evidence conditions match the query's recall frame.

The final evidence is then formatted with structured context rather than flattened into plain text. Each memory is presented with its content and episodic coordinates, including session span, event time when available, location, persons, entities, and topic. These fields expose the conditions under which each memory was observed, allowing the generator to use both the remembered content and its evidential context. The generator is instructed to answer concisely from the provided evidence and to return a structured response from which the final answer can be extracted. Since each retrieved memory remains linked to its identifier and episodic coordinates, RaMem can record which memories entered the generator, where the ground-truth memory was ranked, and whether contextual activation was triggered. These diagnostics help separate failures across memory construction, retrieval, contextual activation, and answer generation.

\begin{table*}[t]
\centering
\caption{Performance on the LoCoMo benchmark. RaMem consistently improves answer quality across different backbone models and question categories.}
\label{tab:locomo_full_results}
\resizebox{\textwidth}{!}{
\begin{tabular}{llcccccccccc}
\toprule
\multirow{2}{*}{\textbf{Model}} & \multirow{2}{*}{\textbf{Method}}
& \multicolumn{2}{c}{\textbf{MultiHop}}
& \multicolumn{2}{c}{\textbf{Temporal}}
& \multicolumn{2}{c}{\textbf{OpenDomain}}
& \multicolumn{2}{c}{\textbf{SingleHop}}
& \multicolumn{2}{c}{\textbf{Average}} \\
\cmidrule(lr){3-4}
\cmidrule(lr){5-6}
\cmidrule(lr){7-8}
\cmidrule(lr){9-10}
\cmidrule(lr){11-12}
& & \textbf{F1} & \textbf{BLEU}
& \textbf{F1} & \textbf{BLEU}
& \textbf{F1} & \textbf{BLEU}
& \textbf{F1} & \textbf{BLEU}
& \textbf{F1} & \textbf{BLEU} \\
\midrule

\multirow{9}{*}{\textbf{GPT-4o}}
& LoCoMo     & 28.00 & 18.47 & 9.09  & 5.78  & 16.47 & 14.80 & \textbf{61.56} & \textbf{54.19} & 28.78 & 23.31 \\
& ReadAgent  & 14.61 & 9.95  & 4.16  & 3.19  & 8.84  & 8.37  & 12.46 & 10.29 & 10.02 & 7.95 \\
& MemoryBank & 6.49  & 4.69  & 2.47  & 2.43  & 6.43  & 5.30  & 8.28  & 7.10  & 5.92  & 4.88 \\
& MemGPT     & 30.36 & 22.83 & 17.29 & 13.18 & 12.24 & 11.87 & 40.16 & 36.35 & 25.01 & 21.06 \\
& A-Mem      & 32.86 & 23.76 & 39.41 & 31.23 & 17.10 & 15.84 & 44.43 & 38.97 & 33.45 & 27.45 \\
& LightMem   & 28.15 & 21.83 & 36.53 & 29.12 & 13.38 & 11.54 & 33.76 & 28.02 & 27.96 & 22.63 \\
& Mem0       & 35.13 & 27.56 & 52.38 & \underline{44.15} & 17.73 & 15.92 & 39.12 & 35.43 & 36.09 & \underline{30.77} \\
& SimpleMem
             & \underline{35.89} & \underline{32.83}
             & \underline{56.71} & 20.57
             & \underline{18.23} & \underline{16.34}
             & 45.41 & 39.25
             & \underline{39.06} & 27.25 \\
& RaMem (Ours)
             & \textbf{47.31} & \textbf{39.08}
             & \textbf{60.09} & \textbf{45.35}
             & \textbf{25.58} & \textbf{20.47}
             & \underline{52.87} & \underline{47.08}
             & \textbf{51.66} & \textbf{43.60} \\

\midrule

\multirow{9}{*}{\textbf{GPT-4.1-mini}}
& LoCoMo     & 25.02 & 21.62 & 12.04 & 10.63 & 19.05 & 17.07 & 18.68 & 15.87 & 18.70 & 16.30 \\
& ReadAgent  & 6.48  & 5.60  & 5.31  & 4.23  & 7.66  & 6.62  & 9.18  & 7.91  & 7.16  & 6.09 \\
& MemoryBank & 5.00  & 4.68  & 5.94  & 4.78  & 5.16  & 4.52  & 5.72  & 4.86  & 5.46  & 4.71 \\
& MemGPT     & 17.72 & 16.02 & 19.44 & 16.54 & 11.29 & 10.18 & 25.59 & 24.25 & 18.51 & 16.75 \\
& A-Mem      & 25.06 & 17.32 & 51.01 & 44.75 & 13.22 & 14.75 & 41.02 & 36.99 & 32.58 & 28.45 \\
& LightMem   & 24.96 & 21.66 & 20.55 & 18.39 & 19.21 & 17.68 & 33.79 & 29.66 & 24.63 & 21.85 \\
& Mem0       & 30.14 & 27.62 & 48.91 & 44.82 & 16.43 & 14.94 & 41.30 & 36.17 & 34.20 & 30.89 \\
& SimpleMem
             & \underline{43.46} & \textbf{38.82}
             & \underline{58.62} & \textbf{50.10}
             & \underline{19.76} & \underline{18.04}
             & \underline{51.12} & \underline{43.53}
             & \underline{43.24} & \underline{37.62} \\
& RaMem (Ours)
             & \textbf{43.73} & \underline{35.99}
             & \textbf{61.57} & \underline{47.02}
             & \textbf{26.35} & \textbf{21.73}
             & \textbf{58.14} & \textbf{51.48}
             & \textbf{54.23} & \textbf{45.86} \\

\midrule

\multirow{9}{*}{\textbf{Qwen3-8b}}
& LoCoMo     & 13.50 & 9.20  & 6.80  & 5.50  & 10.10 & 8.80  & 14.50 & 11.20 & 11.23 & 8.68 \\
& ReadAgent  & 7.20  & 5.10  & 3.50  & 3.10  & 5.50  & 5.40  & 8.10  & 6.20  & 6.08  & 4.95 \\
& MemoryBank & 9.50  & 7.10  & 3.80  & 2.50  & 7.50  & 6.50  & 9.20  & 7.50  & 7.50  & 5.90 \\
& MemGPT     & 14.20 & 9.80  & 5.50  & 4.20  & 12.50 & 10.80 & 11.50 & 9.10  & 10.93 & 8.48 \\
& A-Mem      & 20.50 & 13.80 & 22.50 & 18.20 & 13.20 & 10.50 & 26.80 & 21.50 & 20.75 & 16.00 \\
& LightMem   & 18.53 & 14.23 & 26.78 & 21.52 & 14.12 & 11.24 & 29.48 & 23.83 & 22.23 & 17.71 \\
& Mem0       & 22.42 & 16.83 & 32.48 & 26.13 & 15.23 & 12.54 & 33.05 & 27.24 & 25.80 & 20.69 \\
& SimpleMem
             & \underline{28.97} & \underline{24.93}
             & \underline{42.85} & \textbf{36.49}
             & \underline{15.35} & \textbf{13.90}
             & \underline{46.62} & \underline{40.69}
             & \underline{33.45} & \underline{29.00} \\
& RaMem (Ours)
             & \textbf{36.10} & \textbf{28.07}
             & \textbf{44.17} & \underline{32.72}
             & \textbf{17.71} & \underline{13.79}
             & \textbf{50.59} & \textbf{44.88}
             & \textbf{44.55} & \textbf{37.33} \\

\midrule

\multirow{9}{*}{\textbf{Qwen2.5-3b}}
& LoCoMo     & 4.61  & 4.29  & 3.11  & 2.71  & 4.55  & 5.97  & 7.03  & 5.69  & 4.83  & 4.67 \\
& ReadAgent  & 2.47  & 1.78  & 3.01  & 3.01  & 5.57  & 5.22  & 3.25  & 2.51  & 3.58  & 3.13 \\
& MemoryBank & 3.60  & 3.39  & 1.72  & 1.97  & 6.63  & 6.58  & 4.11  & 3.32  & 4.02  & 3.82 \\
& MemGPT     & 5.07  & 4.31  & 2.94  & 2.95  & 7.04  & 7.10  & 7.26  & 5.52  & 5.58  & 4.97 \\
& A-Mem      & 12.57 & 9.01  & \textbf{27.59} & \textbf{25.07} & 7.12  & 7.28  & 17.23 & 13.12 & 16.13 & 13.62 \\
& LightMem   & 16.43 & 11.39 & 6.92  & 4.56  & 8.06  & 7.23  & 18.28 & 15.24 & 12.42 & 9.61 \\
& Mem0       & 16.89 & 11.54 & 8.52  & 6.23  & 10.24 & 8.82 & 16.47 & 12.43 & 13.03 & 9.76 \\
& SimpleMem
             & \underline{17.03} & \underline{11.87}
             & 21.47 & \underline{19.50}
             & \underline{12.52} & \underline{10.19}
             & \underline{20.90} & \underline{18.01}
             & \underline{17.98} & \underline{14.89} \\
& RaMem (Ours)
             & \textbf{21.74} & \textbf{15.17}
             & \underline{24.45} & 17.00
             & \textbf{13.68} & \textbf{11.41}
             & \textbf{26.95} & \textbf{22.72}
             & \textbf{24.65} & \textbf{19.44} \\

\bottomrule
\end{tabular}
}
\end{table*}

\section{Experiments}
\label{sec:experiments}
We evaluate whether RaMem improves long-term agent memory by making retrieved memories contextually verifiable as evidence. Our experiments follow the SimpleMem~\citep{liu2026simplemem} protocol on LoCoMo~\citep{maharana2024evaluating} and LongMemEval~\citep{wu2024longmemeval}. We evaluate across four backbones, GPT-4o, GPT-4.1-mini, Qwen2.5-3B, and Qwen3-8B, and report token-level F1 and BLEU-1 as the primary metrics. Full details and implementation are provided in Appendix~\ref{app:experimental_setup}.

\subsection{Main Results on Long-Term Memory}
\label{subsec:main_results}

Table~\ref{tab:locomo_full_results} reports the main results on LoCoMo. RaMem achieves the best average F1 across all four backbones and consistently improves over SimpleMem, the strongest structured-memory baseline. The gains are substantial across model families: average F1 increases from 39.06 to 51.66 on GPT-4o, from 43.24 to 54.23 on GPT-4.1-mini, from 33.45 to 44.55 on Qwen3-8B, and from 17.98 to 24.65 on Qwen2.5-3B. Average BLEU also improves across all backbones. Since SimpleMem already uses compressed memory units, multi-view indexing, and intent-aware retrieval, these improvements suggest that RaMem provides a complementary benefit: it improves whether retrieved memories are contextually valid evidence for the current query, rather than merely making memories compact and retrievable. The category-level results further show that RaMem is not simply a temporal retrieval heuristic. Although temporal grounding is a central source of contextual verification, large gains also appear in MultiHop, OpenDomain, and SingleHop questions. For example, on GPT-4o, RaMem improves MultiHop, OpenDomain, and SingleHop F1 by 11.42, 7.35, and 7.46 points, respectively; on GPT-4.1-mini, the strongest gains appear in OpenDomain and SingleHop questions. Similar improvements in MultiHop and SingleHop performance also hold for Qwen3-8B and Qwen2.5-3B. This pattern supports the intended role of contextual reinstatement: time and session information are not used only to answer temporal questions, but to identify which episode or event instance should serve as evidence. A more detailed analysis of the main results is provided in Appendix~\ref{app:main_results_analysis}.

\subsection{Evidence Retrieval Diagnostics}
\label{subsec:retrieval_diagnostics}

Final answer quality alone cannot show whether a memory system retrieves the right evidence, so we evaluate retrieval diagnostics using verified ground-truth memory entries. For each question, we record whether the answer-supporting memory appears in the retrieved set and where it is ranked. Table~\ref{tab:retrieval_diagnostics} shows that RaMem consistently improves evidence retrieval across backbones. On GPT-4.1-mini, Recall@10 increases from 0.7221 to 0.7890, MRR improves from 0.4908 to 0.5476, and capped average rank drops from 8.5292 to 5.9844; Qwen2.5-3B and Qwen3-8B show similar Recall@10 gains of 6.54 and 7.34 points. The gains are even stronger when contextual cues are activated: on the temporal-triggered subset, GPT-4.1-mini improves Recall@10 from 0.6326 to 0.8140, while Qwen2.5-3B improves from 0.5256 to 0.6558. These results show that RaMem retrieves the ground-truth memory more often and ranks it earlier, especially when the query points to a specific episode where related memories can become strong distractors. Generator failures can still occur after the correct memory is retrieved, but RaMem reduces this failure rate across all three backbones, suggesting that contextual reinstatement improves both memory access and the quality of evidence passed to the generator.

\begin{table*}[t]
\centering
\small
\caption{Retrieval diagnostics on LoCoMo. RaMem improves both ground-truth memory recall and ranking across backbone models. Lower values are better for CappedRank@10, HitOnlyRank, and GenFailGivenRetrieved.}
\label{tab:retrieval_diagnostics}
\resizebox{\textwidth}{!}{
\begin{tabular}{llcccccccc}
\toprule
\textbf{Backbone} & \textbf{Method} & \textbf{R@1} & \textbf{R@3} & \textbf{R@5} & \textbf{R@10} & \textbf{MRR} & \textbf{CappedRank@10} & \textbf{HitOnlyRank} & \textbf{GenFailGivenRetrieved} \\
\midrule
\multirow{2}{*}{GPT-4.1-mini}
& SimpleMem & 0.3662 & 0.5636 & 0.6364 & 0.7221 & 0.4908 & 8.5292 & 8.3831 & 0.8336 \\
& Ours      & \textbf{0.4175} & \textbf{0.6305} & \textbf{0.6961} & \textbf{0.7890} & \textbf{0.5476} & \textbf{5.9844} & \textbf{5.6510} & \textbf{0.8109} \\
\midrule
\multirow{2}{*}{Qwen2.5-3B}
& SimpleMem & 0.2740 & 0.4539 & 0.5234 & 0.6247 & 0.3968 & 10.0006 & 9.8928 & 0.9230 \\
& Ours      & \textbf{0.3054} & \textbf{0.4776} & \textbf{0.5809} & \textbf{0.6901} & \textbf{0.4313} & \textbf{8.0143} & \textbf{7.6847} & \textbf{0.9113} \\
\midrule
\multirow{2}{*}{Qwen3-8B}
& SimpleMem & 0.2870 & 0.4903 & 0.5526 & 0.6617 & 0.4178 & 9.7390 & 9.6429 & 0.8693 \\
& Ours      & \textbf{0.3292} & \textbf{0.5435} & \textbf{0.6266} & \textbf{0.7351} & \textbf{0.4669} & \textbf{7.2532} & \textbf{6.9594} & \textbf{0.8340} \\
\bottomrule
\end{tabular}
}
\end{table*}

\subsection{Context Collapse Analysis}
\label{subsec:context_collapse}

\begin{figure*}[t]
\centering
\includegraphics[width=\textwidth,height=4cm]{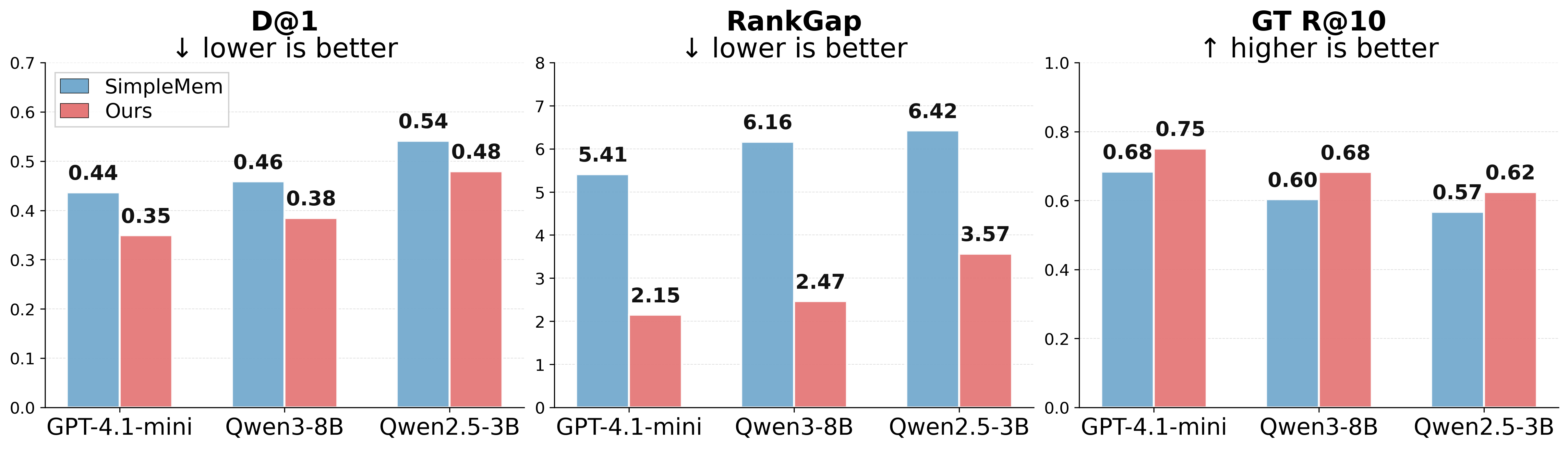}
\caption{\textbf{Context collapse mitigation across backbones}. Each subplot corresponds to one backbone and compares SimpleMem with our method on three diagnostic metrics: D@1, RankGap, and GT R@10. Lower is better for D@1 and RankGap, while higher is better for GT R@10. }
\label{fig:context_collapse}
\end{figure*}

To directly test whether RaMem mitigates context collapse, we analyze \textbf{context distractors}: retrieved memories that are related to the query or ground-truth memory but belong to an incompatible episodic context. We identify distractors using person, entity, topic, and keyword overlap for content relatedness, together with strict temporal or session mismatch for contextual incompatibility. We further define a \textbf{context-confusable subset}, where SimpleMem retrieves at least one strict context distractor in its top-5 results, isolating cases where relevance signals are most likely to select plausible but invalid evidence. Figure~\ref{fig:context_collapse} and Table~\ref{tab:context_collapse_main} show that context collapse is frequent rather than incidental: SimpleMem ranks a context distractor first in 43.66\%, 45.88\%, and 54.14\% of examples on GPT-4.1-mini, Qwen3-8B, and Qwen2.5-3B, respectively. These distractors are related fragments that share entities, topics, or keywords with the query but fail its evidence conditions. RaMem consistently reduces this interference, lowering D@1 to 0.3495, 0.3846, and 0.4796 across the three backbones. It also reduces RankGap by more than three points on GPT-4.1-mini and Qwen3-8B and by nearly three points on Qwen2.5-3B, while improving GT R@10 across all backbones. These results show that RaMem suppresses context-invalid memories without losing the answer-supporting memory, directly supporting our claim that contextual reinstatement distinguishes valid evidence from related but invalid fragments.

\subsection{Context Shuffle Test}
\label{subsec:context_shuffle}

To verify that contextual reinstatement depends on correct content-context binding, we conduct a context shuffle test: memory content is kept unchanged, while episodic context fields are shuffled across entries within the same memory bank. This preserves the amount and format of metadata but breaks the correspondence between a memory and the context in which it was encoded. We focus on temporal/session and full-context shuffling in the main text, with complete results in Appendix~\ref{app:context_shuffle}. Table~\ref{tab:context_shuffle_main} shows that corrupted temporal context consistently hurts retrieval and answer quality. On all verified questions, temporal shuffling reduces GPT-4.1-mini F1 from 0.5291 to 0.5004 and GT R@10 from 0.7550 to 0.6744; on Qwen3-8B, F1 drops from 0.4522 to 0.4287 and GT R@10 from 0.6925 to 0.6204. The effect is much sharper on temporal-triggered questions: GT R@10 drops from 0.8066 to 0.2358 on GPT-4.1-mini and from 0.7170 to 0.1840 on Qwen3-8B, while D@1 rises from 0.2642 to 0.8160 and from 0.2736 to 0.8726, respectively. Full-context shuffling follows the same pattern, suggesting that the dominant failure comes from corrupted session-level temporal grounding. These results show that contextual metadata is not decorative prompt information; it must remain correctly bound to memory content to guide retrieval toward the correct episode.

\begin{table*}[t]
\centering
\small
\caption{Context shuffle test. Memory content is kept unchanged, while selected context fields are shuffled across memory entries. We report both all verified questions and the temporal-triggered subset. Lower is better for D@1 and RankGap.}
\label{tab:context_shuffle_main}
\resizebox{\textwidth}{!}{
\begin{tabular}{lllccccc}
\toprule
\textbf{Backbone} & \textbf{Subset} & \textbf{Variant} 
& \textbf{F1} $\uparrow$ 
& \textbf{GT R@10} $\uparrow$ 
& \textbf{MRR} $\uparrow$ 
& \textbf{D@1} $\downarrow$ 
& \textbf{RankGap} $\downarrow$ \\
\midrule

\multirow{6}{*}{GPT-4.1-mini}
& \multirow{3}{*}{All Verified GT}
& Original & \textbf{0.5291} & \textbf{0.7550} & \textbf{0.5323} & \textbf{0.3428} & \textbf{2.0278} \\
& & Shuffle temporal & 0.5004 & 0.6744 & 0.4871 & 0.5753 & 5.2974 \\
& & Shuffle full context & 0.5079 & 0.6638 & 0.4822 & 0.5872 & 5.3965 \\
\cmidrule(lr){2-8}
& \multirow{3}{*}{Temporal-triggered}
& Original & \textbf{0.6009} & \textbf{0.8066} & \textbf{0.5249} & \textbf{0.2642} & \textbf{1.0404} \\
& & Shuffle temporal & 0.4985 & 0.2358 & 0.1796 & 0.8160 & 21.5029 \\
& & Shuffle full context & 0.5180 & 0.2358 & 0.1762 & 0.8302 & 22.0116 \\

\midrule

\multirow{6}{*}{Qwen3-8B}
& \multirow{3}{*}{All Verified GT}
& Original & \textbf{0.4522} & \textbf{0.6925} & \textbf{0.4502} & \textbf{0.3702} & \textbf{2.1427} \\
& & Shuffle temporal & 0.4287 & 0.6204 & 0.4116 & 0.6518 & 5.7768 \\
& & Shuffle full context & 0.4347 & 0.6157 & 0.4053 & 0.6631 & 5.7634 \\
\cmidrule(lr){2-8}
& \multirow{3}{*}{Temporal-triggered}
& Original & \textbf{0.5317} & \textbf{0.7170} & \textbf{0.4271} & \textbf{0.2736} & \textbf{1.2350} \\
& & Shuffle temporal & 0.4081 & 0.1840 & 0.1250 & 0.8726 & 22.9595 \\
& & Shuffle full context & 0.4134 & 0.1840 & 0.1242 & 0.8868 & 23.0952 \\

\bottomrule
\end{tabular}
}
\end{table*}

\subsection{Hyper-parameter Analysis}
\label{subsec:analysis}

\begin{figure*}[t]
    \centering
\includegraphics[width=0.95\textwidth,height=0.3\textheight,keepaspectratio]{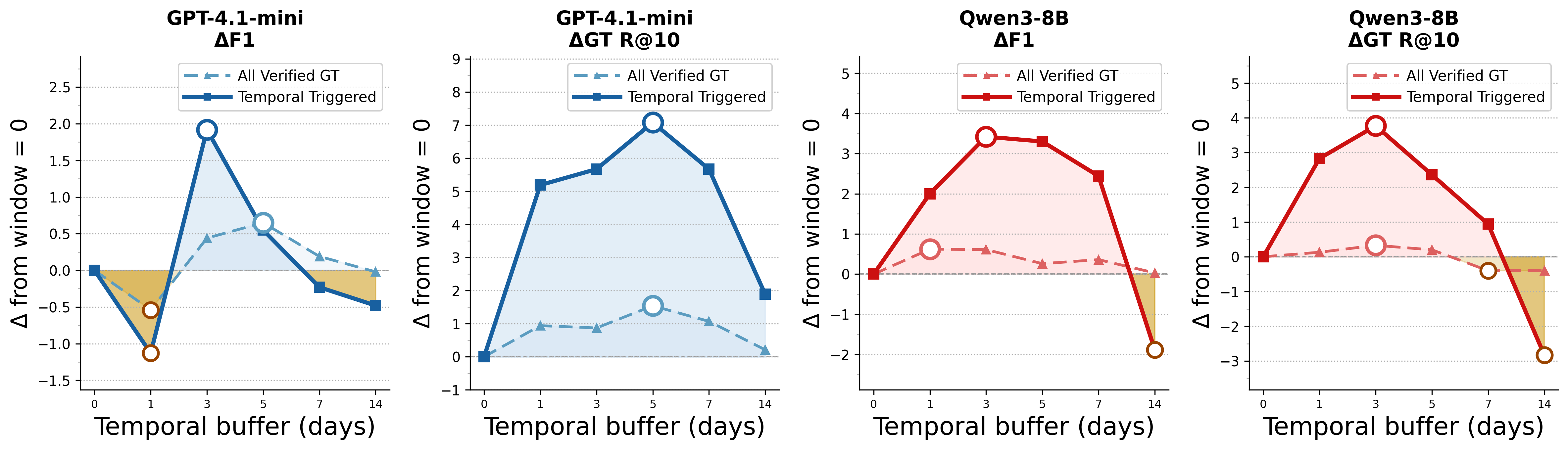}
    \caption{Sensitivity to the temporal reinstatement window. Each subplot shows the effect of varying the temporal buffer on F1 and GT R@10.}
    \label{fig:buffer_sensitivity}
\end{figure*}

We first study the sensitivity of RaMem to the temporal reinstatement window, which controls how much mismatch is allowed between event time and the time at which the event entered memory. As shown in Figure~\ref{fig:buffer_sensitivity}, a moderate buffer provides the best trade-off across GPT-4.1-mini and Qwen3-8B. Very narrow windows can miss valid memories whose mention time is slightly shifted, while overly broad windows admit more context-invalid memories and weaken contextual specificity. This supports our design intuition that temporal context should act as a calibrated episodic coordinate rather than a hard equality constraint or an unrestricted expansion window. Full results with additional diagnostics are provided in  Appendix~\ref{app:buffer_sensitivity}. We also evaluate sensitivity to the retrieval budget in Figure~\ref{fig:budget_sensitivity}, which controls how many memories are passed to the generator. RaMem consistently improves F1 and budget-aligned GT R@$K$ across budgets, and notably matches or exceeds SimpleMem with half the retrieval budget ($K=10$ vs.\ $K=20$) on both GPT-4.1-mini and Qwen3-8B, indicating better evidence ordering and context efficiency; full results are provided in Appendix~\ref{app:budget_sensitivity}.

\subsection{Component Analysis}
\label{subsec:component_analysis}

We analyze which components make retrieved memories contextually verifiable. Since session-level evidence conditions and validity-aware ranking are most directly exercised on temporally grounded queries, we report the temporal-triggered subset in the main text and provide full ablations in Appendix~\ref{app:ablation}. Table~\ref{tab:component_analysis_main} shows that session grounding and validity-aware ranking are the main retrieval-side drivers. Without session context, GT R@10 drops from 0.8066 to 0.5708 on GPT-4.1-mini and from 0.7217 to 0.4245 on Qwen3-8B; D@1 roughly doubles and RankGap increases substantially. Removing context-aware ranking causes a similar degradation, indicating that RaMem must both retrieve session-compatible candidates and order them by recall-condition compatibility. In contrast, removing context-preserved generation leaves retrieval metrics unchanged by design, but reduces F1 from 0.5957 to 0.5626 on GPT-4.1-mini and from 0.5226 to 0.4248 on Qwen3-8B. This shows that even correctly retrieved evidence must retain its structured context for generator to use it.

\begin{table*}[t]
\centering
\small
\caption{Component analysis on the temporal-triggered subset. We report the most diagnostic ablations in the main text. Lower is better for D@1 and RankGap. }
\label{tab:component_analysis_main}
\resizebox{\textwidth}{!}{
\begin{tabular}{llccccc}
\toprule
\textbf{Backbone} & \textbf{Variant}
& \textbf{F1} $\uparrow$
& \textbf{BLEU1} $\uparrow$
& \textbf{GT R@10} $\uparrow$
& \textbf{D@1} $\downarrow$
& \textbf{RankGap} $\downarrow$ \\
\midrule
\multirow{4}{*}{GPT-4.1-mini}
& Full & \textbf{0.5957} & \textbf{0.5319} & \textbf{0.8066} & \textbf{0.2736} & \textbf{1.0354} \\
& w/o Session Context & 0.5596 & 0.4922 & 0.5708 & 0.5189 & 10.0492 \\
& w/o Context-Aware Ranking & 0.5693 & 0.5018 & 0.5708 & 0.5283 & 12.6332 \\
& w/o Context-Preserved Generation & 0.5626 & 0.4968 & 0.8066 & 0.2736 & 1.0354 \\
\midrule
\multirow{4}{*}{Qwen3-8B}
& Full & \textbf{0.5226} & \textbf{0.4695} & \textbf{0.7217} & \textbf{0.2783} & \textbf{1.1685} \\
& w/o Session Context & 0.4578 & 0.4097 & 0.4245 & 0.5849 & 11.9565 \\
& w/o Context-Aware Ranking & 0.4642 & 0.4162 & 0.4434 & 0.5755 & 14.7735 \\
& w/o Context-Preserved Generation & 0.4248 & 0.3567 & 0.7217 & 0.2783 & 1.1685 \\
\bottomrule
\end{tabular}
}
\end{table*}

\subsection{Efficiency Analysis \& Qualitative Case Studies}
\label{subsec:efficiency_cost}

RaMem improves context efficiency without requiring a larger evidence budget. As shown in Table~\ref{tab:budget_efficiency_main}, RaMem with $K=10$ matches or exceeds SimpleMem with $K=20$ on both GPT-4.1-mini and Qwen3-8B, achieving better recall with roughly half the context tokens. 
Full online-cost measurements are provided in Appendix~\ref{app:efficiency_cost}. \textbf{Case studies} in Appendix~\ref{app:case_studies} further show the same mechanism: SimpleMem often retrieves related but context-invalid memories, while RaMem promotes memories whose episodic coordinates satisfy the query's recall frame.

\section{Conclusion}
\label{sec:conclusion}

In this paper, we presented RaMem, a contextual reinstatement framework for long-term agent memory. RaMem makes retrieved memories verifiable by recovering the episodic conditions under which they should be used and by carrying this context into answer generation. Experiments across multiple backbones on LoCoMo show consistent gains in answer quality, ground-truth memory retrieval, and robustness against context-invalid distractors. These results suggest that effective long-term agent memory requires not only making past information available, but also ensuring that retrieved memories are valid under the current query context.

\bibliographystyle{unsrtnat}
\bibliography{citation}


\appendix
\section*{Appendix}

\section{Limitations and Broader Impact}
\label{sec:limitations_impact}

\subsection{Limitations}
This work studies contextual reinstatement for long-term agent memory under textual interaction histories. The current instantiation primarily relies on session-level temporal grounding, participant and entity cues, and structured memory fields to determine whether retrieved memories are contextually valid evidence. Richer forms of context, such as causal dependencies, affective states, evolving task goals, or implicit social relations, may require more expressive memory representations and grounding mechanisms. The framework also depends on the quality of upstream memory extraction and query analysis. Missing episodic conditions in the memory bank, or incorrectly grounded recall conditions from the query, can still limit retrieval quality. In addition, contextual verification introduces design choices about how strictly recall conditions should constrain retrieval and how much temporal mismatch should be tolerated. Finally, our experiments focus on established long-term memory benchmarks, and extending contextual reinstatement to multimodal memories, tool-use trajectories, and real-time deployed agents remains an important direction for future work.

\subsection{Broader Impact}
More reliable long-term agent memory can improve personalized assistants, long-horizon task support, and other interactive systems that depend on accurate recall of prior experiences. By emphasizing contextual validity, the proposed approach may also reduce errors caused by using memories from the wrong episode or outdated user state. At the same time, stronger memory mechanisms require careful deployment. Systems that store and retrieve long-term user information should include clear consent, retention controls, deletion mechanisms, and safeguards against exposing sensitive information or over-relying on stale memories. These considerations are especially important when memory is used to personalize decisions or support user-facing recommendations.

\section{Related Work}
\label{app:related_work}

\subsection{LLM Agents and Long-Term Memory}
\label{subsec:rw_memory_systems}

The rapid development of large language models has been driven by advances in large-scale pretraining, which have enabled strong general-purpose capabilities across understanding~\citep{chang2025survey,li2025climatellm}, reasoning~\citep{chen2026self,chen2025tourrank,xia2026trackrec}, coding~\citep{ping2026coevo,ping2026poet}, recommendation~\citep{gu2025r,zhao2025hierarchical}, and other domains~\citep{ping2025hdlcore,weng2026temporalbench,ye2026ts}. Building on these capabilities, recent work has increasingly studied LLM-based agents that can interact with external tools, environments, users, and other agents to solve complex tasks beyond single-turn text generation~\citep{yang2026tournament}. In particular, multi-agent LLM systems~\citep{chen2026unitymas,yang2026learning,yang2025toward} organize multiple specialized agents into collaborative workflows~\citep{yang2026adaptive,yang2025maestro}, and have been widely explored in settings such as question answering~\citep{zhang2026oases}, decision making~\citep{yang2026auditing}, and other applications~\citep{li2026fortis,ping2025verimoa,ye2025llm}. As these agentic systems move from isolated tasks to long-horizon interactions, persistent memory becomes a key mechanism for maintaining continuity, reusing prior experience, and grounding future decisions in past context.

Long-term memory has become an important mechanism for extending LLM agents beyond the limits of a single context window~\citep{202603.0359,10.1145/3748302, hu2025memory,li-etal-2025-treble,li2026defensespromptattackslearn,du2026memoryautonomousllmagentsmechanisms,qin2026memory}. A natural early direction is to maintain external memory buffers or virtual context systems, where past interactions can be paged, summarized, or retrieved when the model needs them ~\citep{park2023generativeagentsinteractivesimulacra, shinn2023reflexion}. MemGPT~\citep{packer2024memgptllmsoperatingsystems} frames this problem through an operating-system analogy, allowing agents to move information between active context and external storage. MemoryBank~\citep{zhong2024memorybank} stores and updates long-term user-related memories for personalized interaction. SCM~\citep{wang2023enhancing} and MemoryOS~\citep{kang2025memory} further explore controller-style memory streams and modular memory management. These systems establish the basic architecture of non-parametric agent memory, but they often keep memory close to raw dialogue records or summaries, which can preserve redundancy and leave referential or temporal ambiguity unresolved.

A second line of work introduces more structured memory representations~\citep{du2025rethinkingmemoryllmbased,yang2026graphbasedagentmemorytaxonomy}. Zep~\citep{rasmussen2025zep} represents evolving conversational information with a temporal knowledge graph, while A-Mem~\citep{xu2025mem} organizes memory in a self-evolving, note-linking structure inspired by Zettelkasten. Mem0~\citep{chhikara2025mem0} extracts compact persistent facts from conversations and further extends them with graph-based memory to model relational structure. LightMem~\citep{fang2025lightmem} pursues a lightweight architecture inspired by classic memory models, and O-Mem~\citep{wang2025mem} studies broader omni-memory organization for long-horizon agents. These methods move beyond passive context retention by adding structure to memory storage and update. Their main focus, however, remains on how to represent, consolidate, and maintain useful information over time.

More recent work has shifted from memory storage toward memory distillation and adaptive management~\citep{xu2025mem, tan2025prospect, li2026autonomytaxdefensetraining,zhang2026adaptive}. SimpleMem~\citep{liu2026simplemem} compresses dialogue into self-contained memory units, indexes them through semantic, lexical, and symbolic views, and uses intent-aware retrieval to reduce token cost. NEMORI~\citep{nan2025nemori} asks what deserves to be stored by relating retention to predictability and prediction error. AgeMem~\citep{yu2026agentic} exposes long-term and short-term memory operations as actions and optimizes memory behavior through reinforcement learning. MemAgent~\citep{yu2025memagent} also treats memory update as a trainable process for long-context reading. Together, these works show a clear progression from storing history, to structuring memory, to learning or adapting memory operations. Our work follows this progression but studies a different point in the memory pipeline: after memories have been stored, retrieval can still fail when semantically similar memories from different episodes become difficult to distinguish~\citep{maharana2024evaluating}.

\subsection{Context Management and Agentic Retrieval}
\label{subsec:rw_context_retrieval}

The problem of using long histories is closely related to long-context modeling and retrieval-augmented generation    ~\citep{beltagy2020longformer,Li_Ji_Wu_Li_Qin_Wei_Zimmermann_2024,Li_2025_CVPR, zaheer2020big, li-etal-2024-retrieval}. Long-context LLMs and position-extrapolation methods attempt to expand the amount of text that can be directly processed by the model~\citep{chen2023extending, peng2026yarnefficientcontextwindow}, while sparse attention, recurrent architectures, and state-space models seek more efficient sequence processing~\citep{child2019generatinglongsequencessparse,li2025secureondevicevideoood,limm,dai-etal-2019-transformer, gu2024mambalineartimesequencemodeling}. Even with longer contexts, empirical work on the ``lost-in-the-middle'' effect shows that simply placing more information in the prompt does not guarantee reliable use of relevant evidence~\citep{liu2024lost}. Prompt compression methods such as LLMLingua~\citep{jiang2023llmlingua} reduce context cost by shortening inputs, while retrieval-augmented generation decouples knowledge storage from generation by retrieving a smaller evidence set at inference time~\citep{lewis2020retrieval}. Follow-up RAG systems, including Self-RAG~\citep{asai2023self}, active retrieval methods~\citep{jiang2023active}, GraphRAG~\citep{edge2024local}, and LightRAG~\citep{guo2024lightrag}, improve retrieval control, graph-structured evidence aggregation, or query-dependent evidence use.

Although these techniques improve context access, long-term conversational memory introduces a distinct challenge~\citep{huet2025episodicmemoriesgenerationevaluation,pink2025positionepisodicmemorymissing}. Unlike static documents or knowledge bases, dialogue histories contain recurring people, places, topics, and events across multiple sessions~\citep{maharana2024evaluating, jia-etal-2025-evaluating}. A retrieved memory may therefore be highly relevant in wording but still refer to the wrong episode~\citep{liu2025echolargelanguagemodel, shu2026rememreasoningepisodicmemory}. This issue is only partially addressed by standard dense retrieval, lexical retrieval, or symbolic filtering\citep{karpukhin2020densepassageretrievalopendomain, 10.1561/1500000019}. ReadAgent~\citep{lee2024human} compresses very long contexts into gist memories, which improves scalability but does not directly resolve episode-level ambiguity. SimpleMem~\citep{liu2026simplemem} uses multi-view retrieval to combine semantic, lexical, and symbolic access paths, but its retrieval objective is still primarily organized around retrieving compact relevant memories under a token budget.

A parallel trend makes retrieval itself more agentic~\citep{singh2026agenticretrievalaugmentedgenerationsurvey}. ReAct~\citep{yao2022react} combines reasoning and acting, Reflexion~\citep{shinn2023reflexion} stores verbal feedback from past trials, and Self-RAG~\citep{asai2023self} lets the model decide when retrieval or critique is needed. More recent agent-training frameworks such as Search-R1~\citep{jin2025searchr1trainingllmsreason}, Agent-R1~\citep{cheng2025agentr1trainingpowerfulllm}, and RAGEN~\citep{wang2025ragenunderstandingselfevolutionllm} extend this idea to multi-turn retrieval or tool-use settings. For long-term conversational memory, MemR$^3$~\citep{du2025memr} introduces a controller that routes among retrieve, reflect, and answer actions while maintaining an explicit evidence-gap state. These methods address when to retrieve more evidence and how to control multi-step evidence collection. Our work is complementary: instead of emphasizing retrieval depth or closed-loop control, we focus on the validity of retrieved memories under the query's episodic context.

\section{Method}
\label{app:implementation_details}

This appendix provides additional details for the proposed contextual reinstatement framework. The main paper describes the method at the level of content-context binding, recall frame induction, selective reinstatement, and context-preserved evidence synthesis. Here we describe the full system pipeline, including memory construction, memory schema, indexing, retrieval, answer generation, and diagnostic logging.

\subsection{End-to-End Pipeline}
\label{app:end_to_end_pipeline}

The system operates in two stages. The first stage is an offline memory construction stage. Given a long conversation sample, the system extracts structured memory units from the dialogue history, associates each memory with episodic context, indexes the resulting memories, and stores a frozen memory database for later evaluation. The second stage is an online question-answering stage. Given a question, the system loads the frozen memory database, induces a recall frame from the question, retrieves contextually relevant memories, generates an answer, and records both final metrics and retrieval diagnostics.

The pipeline is implemented with six main components. The \texttt{LLMClient} provides a unified interface to OpenAI-compatible language models, including hosted models and locally served models. The \texttt{EmbeddingModel} encodes memory contents and query strings for dense retrieval. The \texttt{VectorStore} stores memory entries and supports semantic, lexical, and structured retrieval. The \texttt{MemoryBuilder} converts dialogue windows into structured episodic memory units. The \texttt{HybridRetriever} performs query analysis, optional planning, semantic retrieval, lexical retrieval, structured retrieval, selective contextual reinstatement, and optional reflection. The \texttt{AnswerGenerator} formats retrieved memories with their context fields and produces the final answer.

We separate memory construction from evaluation for reproducibility and efficiency. Once memory construction is finished, each conversation sample has an independent frozen database. All later retrieval, generation, prompt ablations, and diagnostic analyses can reuse the same memory database without rebuilding memories. This separation also avoids cross-sample contamination and makes different retrieval variants directly comparable under the same stored memory bank.

\subsection{Dialogue Normalization and Episodic Memory Construction}
\label{app:memory_construction}

Each raw conversation is first normalized into a sequence of dialogue turns. A dialogue turn contains the speaker, utterance content, optional timestamp, and session-level temporal boundaries. We use the following abstract structure:
\[
d_j =
\big(
\mathrm{id}_j,
\mathrm{speaker}_j,
\mathrm{text}_j,
\tau_j,
s_j^{\mathrm{start}},
s_j^{\mathrm{end}}
\big),
\]
where $\tau_j$ denotes the original timestamp when available, and $[s_j^{\mathrm{start}},s_j^{\mathrm{end}}]$ denotes the session span of the dialogue turn. The session span is later used as a stable episodic coordinate for memory retrieval.

The dialogue history is processed with overlapping sliding windows. Let $W$ denote the window size and $O$ denote the overlap size. Consecutive windows advance by $W-O$ turns. The overlap reduces boundary errors by allowing memories that depend on local context to be extracted even when the relevant evidence lies near a window boundary. In our implementation, memory extraction is parallelized across windows when possible. This enables local model servers to batch multiple extraction requests and improves throughput.

For each window, the language model is prompted to extract a set of structured memory entries. The extraction prompt asks the model to avoid producing a single coarse summary of the entire window. Instead, it should produce multiple self-contained memories, each corresponding to a distinct event, preference, fact, decision, or relation that may be useful later. The expected output is a JSON object with a list of memory entries:
\[
\mathrm{Extract}(w)
\rightarrow
\{
m_1,\ldots,m_R
\}.
\]
Each raw memory entry contains a lossless restatement, keywords, optional timestamp, location, persons, entities, and topic. The restatement is designed to be understandable without the original dialogue window. In practice, the extraction prompt enforces an upper bound on the number of memories per window to avoid over-fragmentation and excessive index size.

After extraction, each memory entry is matched back to a real session span. The language model may extract a timestamp from the content, but this timestamp can be noisy or may refer to the event time rather than the conversational mention time. We therefore use the session spans from the source dialogue window as the primary encoding-time coordinate. If the extracted timestamp falls inside a known session span, the memory is assigned to that span. If it does not fall inside any span, the system assigns the closest session span. This prevents memories from being left temporally ungrounded.

The resulting memory entry has the following abstract schema:
\[
m_i = (x_i, k_i, e_i),
\]
where $x_i$ is the lossless restatement, $k_i$ is a set of lexical keywords, and $e_i$ is an episodic context frame:
\[
e_i =
\{
\tau_i^{\mathrm{event}},
\tau_i^{\mathrm{mention}},
[s_i^{\mathrm{start}},s_i^{\mathrm{end}}],
p_i,
\ell_i,
u_i,
z_i
\}.
\]
Here $\tau_i^{\mathrm{event}}$ denotes the event time inferred from the memory content when available, $\tau_i^{\mathrm{mention}}$ denotes an explicitly mentioned temporal cue when available, $[s_i^{\mathrm{start}},s_i^{\mathrm{end}}]$ denotes the session span, $p_i$ denotes persons or participants, $\ell_i$ denotes location, $u_i$ denotes entities, and $z_i$ denotes the topic.

This schema distinguishes event time from encoding context. The event time describes when the remembered event occurred. The session span describes when the information entered the conversation and was encoded into memory. This distinction is important because many long-term conversational questions refer to when something was discussed or remembered, not only when the real-world event happened.

\subsection{Multi-View Indexing and Frozen Memory Stores}
\label{app:indexing}

Each memory entry is inserted into a vector store with both content and context fields. The stored schema contains an entry identifier, the lossless restatement, keywords, event timestamp, location, persons, entities, topic, session start time, session end time, mention date when available, and dense vector representation. The dense vector is computed from the lossless restatement:
\[
v_i = f_{\mathrm{emb}}(x_i),
\]
where $f_{\mathrm{emb}}$ is the document encoder. At retrieval time, the query encoder maps the query or rewritten query into the same vector space. If the embedding model supports different query and document encoding modes, the system uses the query-specific encoding mode for questions and the document encoding mode for memories.

The memory bank supports three retrieval views. The semantic view performs dense retrieval over memory vectors. The lexical view performs full-text or BM25-style retrieval over lossless restatements and keywords. The structured view filters or scores memories using episodic context fields such as persons, entities, locations, timestamps, and session spans. These views correspond to different access paths into the same memory bank:
\[
\mathcal{R}_{\mathrm{sem}},
\quad
\mathcal{R}_{\mathrm{lex}},
\quad
\mathcal{R}_{\mathrm{ctx}}.
\]
The semantic view provides broad paraphrase-level recall. The lexical view captures exact entity and keyword matches. The structured view enables context-aware discrimination when reliable contextual cues are available.

After construction, each conversation sample is copied into an independent frozen database. During evaluation, the frozen database is loaded in read-only fashion. This design ensures that all questions from the same sample use the same memory bank, while different samples remain isolated. It also allows retrieval and generation ablations to be run repeatedly without changing the underlying memories.

\subsection{Contextual Recall and Selective Retrieval}
\label{app:retrieval_details}

At inference time, RaMem first analyzes the question to infer both the information need and the evidence conditions that a valid memory should satisfy. The analyzer extracts keywords, persons, locations, entities, topic hints, temporal expressions, and a grounded time range when possible. When the global date range of the interaction history is available, it is included in the analysis prompt so that relative expressions such as ``last August'', ``that summer'', or ``the first weekend of May'' can be resolved within the current memory timeline.

We denote the analyzed query as a pair $(r_q,c_q)$, where $r_q$ is the information need and $c_q$ is the contextual recall frame:
\[
c_q =
\{
\hat{\tau}_q,
\hat{p}_q,
\hat{\ell}_q,
\hat{u}_q,
\hat{z}_q
\}.
\]
Here $\hat{\tau}_q$ is the grounded query time range when inferable, $\hat{p}_q$ denotes persons, $\hat{\ell}_q$ denotes locations, $\hat{u}_q$ denotes entities, and $\hat{z}_q$ denotes topic or episode hints. Since induced recall conditions can still be ambiguous, RaMem does not apply them unconditionally; instead, contextual retrieval is activated only when the relevant cues can be grounded reliably.

RaMem first retrieves a broad content-based candidate set. When enabled, an optional LLM-based planning step analyzes the query and generates a small set of targeted semantic search queries while retaining the original query. When this step is disabled, RaMem falls back to semantic retrieval with the original query alone. Dense retrieval ranks memories by embedding similarity between semantic queries and memory content, while lexical retrieval ranks memories by keyword or BM25-style matching. The two ranked lists are combined with reciprocal rank fusion:
\[
S_{\mathrm{content}}(q,m_i)
=
\sum_{r \in \{\mathrm{sem},\mathrm{lex}\}}
\frac{1}{\kappa + \mathrm{rank}_r(q,m_i)},
\]
where $\kappa$ is a smoothing constant and $\mathrm{rank}_r(q,m_i)$ is the rank of memory $m_i$ under retrieval view $r$. This produces a broad set of memories that may contain relevant content, but it does not yet verify whether these memories satisfy the query's evidence conditions.

Contextual reinstatement is activated selectively. For temporal context, the system uses a two-stage activation rule. First, the raw question must contain a detectable temporal cue, such as a date, month, season, relative week, or similar expression. Second, the query analyzer must ground this cue into a valid time range. If either condition fails, RaMem does not impose temporal filtering and instead falls back to content-based retrieval. This prevents spurious or hallucinated time ranges from suppressing the correct memory.

When the recall frame provides grounded conditions, RaMem constructs a context-compatible candidate list rather than directly treating all content-relevant memories as equally valid evidence. Let $\mathcal{C}_{\mathrm{content}}(q)$ be the content-retrieved candidates and let $\mathcal{C}_{\mathrm{ctx}}(q)$ denote memories retrieved under grounded contextual conditions. The candidate list used for evidence synthesis is
\[
\mathcal{C}_q =
\begin{cases}
\mathrm{Dedup}_{\mathrm{first}}\big(
\mathcal{C}_{\mathrm{ctx}}(q) \oplus \mathcal{C}_{\mathrm{content}}(q)
\big), & \text{if recall conditions are grounded},\\
\mathcal{C}_{\mathrm{content}}(q), & \text{otherwise}.
\end{cases}
\]
Here $\oplus$ denotes ordered concatenation, and $\mathrm{Dedup}_{\mathrm{first}}$ keeps the first occurrence of each memory according to its identifier. Thus, grounded context gives priority to memories that are applicable under the query's evidence conditions, while content-relevant candidates are retained as fallback evidence. When the induced context is uncertain, it is not imposed as a hard constraint.

For temporal conditions, contextual retrieval is instantiated with session-overlap compatibility. Given a grounded query range $\hat{\tau}_q=[t_q^{\mathrm{start}},t_q^{\mathrm{end}}]$ and a memory session span $\tau_i^{\mathrm{session}}=[s_i^{\mathrm{start}},s_i^{\mathrm{end}}]$, memory $m_i$ is temporally compatible if
\[
s_i^{\mathrm{start}} \leq t_q^{\mathrm{end}}
\quad \mathrm{and} \quad
s_i^{\mathrm{end}} \geq t_q^{\mathrm{start}}.
\]
A small symmetric buffer can be applied to $\hat{\tau}_q$ before filtering. This buffer tolerates mismatch between when an event happened and when it was mentioned in the interaction history. Within the compatible set, memories are ordered by content relevance and temporal proximity, where proximity prioritizes sessions that overlap the grounded query range and then favors nearby session intervals. In the current implementation, temporal/session compatibility is the most explicit context dimension. Person, entity, and location fields provide additional structured access paths when available, while topic information is preserved for generation-time grounding.

The final retrieved set is obtained by prioritizing context-compatible candidates and then retaining high-ranked content candidates as fallback evidence. If optional reflection is enabled, the system can perform an additional retrieval round by checking whether the current evidence is sufficient and issuing a small number of follow-up semantic searches for missing information.

\subsection{Context-Preserved Answer Generation and Diagnostics}
\label{app:generation_and_diagnostics}

After validity-aware retrieval, RaMem prepares the final evidence list by merging the retrieved candidates while preserving retrieval priority. Let $\mathcal{C}_q^{\mathrm{content}}$ denote candidates from content-based retrieval and let $\mathcal{C}_q^{\mathrm{ctx}}$ denote candidates retained or promoted through contextual reinstatement. RaMem first places context-compatible candidates before general content-relevant fallback evidence and then applies stable deduplication:
\[
\mathcal{L}_q
=
\mathrm{Dedup}_{\mathrm{first}}
\big(
\mathcal{C}_q^{\mathrm{ctx}}
\oplus
\mathcal{C}_q^{\mathrm{content}}
\big),
\]
where $\oplus$ denotes ordered concatenation, and $\mathrm{Dedup}_{\mathrm{first}}$ keeps the first occurrence of each memory according to its identifier. The final evidence list is obtained by taking the first $K$ memories from the deduplicated list:
\[
\mathcal{C}_q^{\mathrm{final}}
=
\mathrm{Head}_K(\mathcal{L}_q).
\]
When contextual reinstatement is not activated, $\mathcal{C}_q^{\mathrm{ctx}}$ is empty and the same operation reduces to content-based evidence selection. If optional reflection is enabled, RaMem may issue a small number of additional semantic searches for missing information before the final merge and deduplication step.

Retrieved memories are not passed to the generator as plain text alone. Each memory is formatted with both its content and episodic coordinates:
\[
\mathrm{Format}(m_i)
=
[
x_i;
\tau_i^{\mathrm{session}};
\tau_i^{\mathrm{event}};
\tau_i^{\mathrm{mention}};
\ell_i;
p_i;
u_i;
z_i
].
\]
In text form, each context block contains the lossless memory restatement, session range, event or mention time when available, location, persons, related entities, and topic. This preserves the evidence conditions used during retrieval and allows the generator to resolve temporal, entity, and topic constraints explicitly.

The generator receives the question and a ranked list of formatted memories. To control context length, the system packs memories in retrieval order until a maximum character budget is reached. The generator is instructed to answer only from the provided evidence and to return a JSON object containing a reasoning field and an answer field:
\[
G(q,\mathcal{C}_q^{\mathrm{final}})
\rightarrow
\{
\mathrm{reasoning},
\mathrm{answer}
\}.
\]
For entity-centric questions, the prompt can request the minimal entity phrase. For yes/no questions, it can request a concise yes/no answer. After generation, the system extracts the answer field and applies light normalization before computing metrics.

The pipeline records detailed diagnostics for each question. Each evaluation record stores the question, reference answer, generated answer, retrieved memory identifiers, number of retrieved memories, ground-truth memory identifier when available, ground-truth rank among retrieved memories, guard status, grounded time window, buffered time window, contextual activation status, and final metrics. The system also exports reconstructed generator contexts by mapping retrieved identifiers back to full memory rows. This makes it possible to inspect exactly what evidence the generator saw.

These diagnostics support failure decomposition. If the ground-truth memory is absent from the memory bank, the error is attributed to memory construction. If it exists but is not retrieved, the error is attributed to retrieval or contextual activation. If it is retrieved but the answer is still wrong, the error is attributed to evidence synthesis or generation. This decomposition is essential because final answer quality alone does not reveal whether RaMem improves access to the correct episode or only changes generator behavior.

\subsection{Local Model Serving and Experimental Reproducibility}
\label{app:serving_reproducibility}

The system supports both hosted models and local models served through an OpenAI-compatible API. For local Qwen-based experiments, a vLLM server provides the LLM interface, and a local embedding model provides dense representations for memory contents and queries. Each model variant is assigned a separate run namespace, so its frozen databases, evaluation files, context exports, and ground-truth verification files do not overwrite those of other variants.

The build and evaluation scripts expose the main configuration options through environment variables. These include the model path, served model name, maximum model length, GPU memory utilization, embedding model path, embedding batch size, number of memory build workers, and number of evaluation workers. During memory construction, the extraction prompt requests a valid JSON object and limits output length. Since smaller local models can produce malformed JSON, the parser uses several recovery strategies, including code-block extraction, bracket balancing, trailing-comma cleanup, partial recovery from complete objects inside the memory array, and retry. These engineering details improve reproducibility across model sizes while preserving the same high-level memory schema.

During evaluation, questions can be processed in parallel. Each worker runs the full retrieve-generate-evaluate pipeline, and retrieved contexts are stored in a thread-local side channel to avoid mixing evidence across questions. The same evaluation code supports both the baseline retrieval mode and the contextual reinstatement mode. In the baseline mode, temporal contextual reinstatement is disabled and the system follows the original hybrid retrieval behavior. In the contextual reinstatement mode, session-overlap retrieval, temporal proximity reranking, guard diagnostics, and context-preserved generation are enabled according to the specified configuration.

\section{Experiments}

\subsection{Experimental Setup}
\label{app:experimental_setup}

\paragraph{Dataset.}
We follow the SimpleMem evaluation protocol and evaluate on LoCoMo~\citep{maharana2024evaluating}, a long-term conversational memory benchmark with multi-session dialogue histories and questions that require recovering information from prior interactions. The benchmark contains recurring entities, interleaved topics, and temporal shifts across sessions, making it suitable for evaluating whether a memory system can identify the correct episode rather than only retrieve related fragments. Following our evaluation split, we use samples from the dataset, and evaluate questions from four categories: multi-hop, temporal, open-domain, and single-hop questions. For analyses that require a verified answer-supporting memory, we use the subset with a reliable single ground-truth memory entry. For LongMemEval~\citep{wu2024longmemeval}, we follow the standard accuracy-style evaluation used in SimpleMem, where a judge model compares the generated answer with the reference answer and labels the response as correct or incorrect based on semantic and temporal consistency.

\paragraph{Baselines and Backbones.}
For broad comparison, we use the same representative memory baselines reported in SimpleMem, including the full-history LoCoMo setting~\citep{maharana2024evaluating}, ReadAgent~\citep{lee2024human}, MemoryBank~\citep{zhong2024memorybank}, MemGPT~\citep{packer2024memgptllmsoperatingsystems}, A-Mem~\citep{xu2025mem}, LightMem~\citep{fang2025lightmem}, Mem0~\citep{chhikara2025mem0}, and SimpleMem~\citep{liu2026simplemem}. We reproduced SimpleMem with its official repository and obtained comparable results, so we report baseline numbers from the SimpleMem paper to maintain consistency with its evaluation protocol. These methods cover full-context prompting, gist-style reading memory, persistent user memory, virtual memory, structured or graph-based memory, lightweight retrieval, and semantic-compression-based memory. Following SimpleMem, we evaluate across both closed-source and open-source backbones: GPT-4o, GPT-4.1-mini, Qwen2.5-3B, and Qwen3-8B. For diagnostic analyses that require retrieved memory traces, we compare RaMem against a SimpleMem-style baseline reproduced inside our current implementation, so that both methods share the same serving stack, memory database, and evaluation scripts.

\paragraph{Metrics.}
For answer quality, we report token-level F1 and BLEU-1, following the LoCoMo evaluation protocol. We also report category-level results for multi-hop, temporal, open-domain, and single-hop questions. To verify whether performance gains come from better memory access, we compute retrieval diagnostics using verified ground-truth memory entries, including Recall@$K$, MRR, capped ground-truth rank, hit-only rank, and generator failure conditioned on the ground-truth memory being retrieved. For context-collapse analysis, we measure the rate of context distractors at top ranks, average distractor load in the top-5, and the rank gap between the ground-truth memory and the strongest context distractor. Unless otherwise stated, retrieval diagnostics use $K_{\max}=10$.

\paragraph{Implementation Details.}
We match the SimpleMem~\citep{liu2026simplemem} implementation as closely as possible. Dialogue histories are segmented into overlapping windows, and memory entries are stored in LanceDB with multi-view indexing. Dense retrieval uses Qwen3-Embedding-0.6B with 1024-dimensional embeddings, lexical retrieval uses BM25-style sparse indexing, and structured retrieval uses metadata fields such as timestamps, persons, entities, locations, and session-level context, while topic information is preserved for generation-time grounding. Retrieval budgets and windowing parameters are fixed within each compared setting so that methods are evaluated under the same memory construction and evidence-selection protocol. RaMem keeps the same memory construction and indexing backbone, but replaces the original retrieval stage with contextual reinstatement: it grounds recall conditions from the query, activates temporal retrieval only when reliable cues are available, applies a default $\pm 5$ day temporal buffer, and preserves structured fields such as session, time, location, persons, entities, and topic in the generation prompt. OpenAI models are accessed through OpenAI-compatible APIs, while Qwen models are served locally with vLLM when applicable. All reported comparisons use the same generator backbone within each setting. For LongMemEval-S, we use a benchmark-specific adapter on top of the core RaMem retrieval pipeline. The adapter preserves session-level memory construction and, when needed, applies exact-match fallback over generated memory entries and raw-haystack fallback over salient prior messages; these adapter-specific fallbacks are used only for LongMemEval-S evaluation and are separate from the core RaMem method.

\paragraph{Main Prompts.}
We provide the main prompts used in RaMem for memory construction, query-side recall condition induction, retrieval planning, and final answer generation. Runtime variables are shown in braces, e.g., \texttt{\{query\}} and \texttt{\{context\_str\}}.

\begin{tcolorbox}[
    title={Memory Extraction Prompt},
    colback=gray!3,
    colframe=gray!60,
    fonttitle=\bfseries,
    breakable
]
\small
\textbf{System.}
You are a professional information extraction assistant, skilled at extracting structured, unambiguous information from conversations. You must output valid JSON format.

\vspace{0.4em}
\textbf{User.}
Extract all valuable information from the following dialogues and convert them into structured memory entries.

\vspace{0.4em}
\textbf{Input.}
\texttt{\{dialogue\_text\}}

\vspace{0.4em}
\textbf{Requirements.}
\begin{enumerate}[leftmargin=1.5em,itemsep=0.15em]
    \item Generate enough memory entries to capture all useful information in the dialogues.
    \item Do not use pronouns or unresolved relative time expressions.
    \item Each \texttt{lossless\_restatement} must be complete, independent, and understandable.
    \item Extract structured fields: \texttt{keywords}, \texttt{timestamp}, \texttt{location}, \texttt{persons}, \texttt{entities}, and \texttt{topic}.
    \item Create separate entries for separate facts, preferences, plans, events, locations, relationships, work details, and media references.
    \item Return valid JSON only.
\end{enumerate}

\vspace{0.4em}
\textbf{Output Format.}
\begin{quote}
\small\ttfamily
\{ \\
\quad "memory\_entries": [ \\
\quad\quad \{ \\
\quad\quad\quad "lossless\_restatement": "Complete unambiguous restatement.", \\
\quad\quad\quad "keywords": ["keyword1", "keyword2"], \\
\quad\quad\quad "timestamp": "YYYY-MM-DDTHH:MM:SS or null", \\
\quad\quad\quad "location": "location name or null", \\
\quad\quad\quad "persons": ["name1", "name2"], \\
\quad\quad\quad "entities": ["entity1", "entity2"], \\
\quad\quad\quad "topic": "topic phrase" \\
\quad\quad \} \\
\quad ] \\
\}
\end{quote}

\textbf{Constraint.}
Return only the JSON object.
\end{tcolorbox}

\begin{tcolorbox}[
    title={Query Analysis Prompt},
    colback=gray!3,
    colframe=gray!60,
    fonttitle=\bfseries,
    breakable
]
\small
\textbf{System.}
You are a query analysis assistant. You must output valid JSON format.

\vspace{0.4em}
\textbf{Optional Temporal Context.}
Conversation context: This conversation took place between \texttt{\{CONV\_DATE\_START\}} and \texttt{\{CONV\_DATE\_END\}}. If the query uses relative or implicit time references, resolve them using this date window. A temporal signal is confirmed in this query. Extract start and end dates for any time period mentioned.

\vspace{0.4em}
\textbf{User.}
Analyze the following query and extract key information.

\vspace{0.4em}
\textbf{Query.}
\texttt{\{query\}}

\vspace{0.4em}
\textbf{Fields to Extract.}
\begin{enumerate}[leftmargin=1.5em,itemsep=0.15em]
    \item \texttt{keywords}: names, places, topic words, and other useful search terms.
    \item \texttt{persons}: person names mentioned.
    \item \texttt{time\_expression}: time expression, if any.
    \item \texttt{time\_range\_start}: ISO start date if a time period is mentioned.
    \item \texttt{time\_range\_end}: ISO end date if a time period is mentioned.
    \item \texttt{location}: location, if any.
    \item \texttt{entities}: companies, products, organizations, or other named entities.
\end{enumerate}

\vspace{0.4em}
\textbf{Output Format.}
\begin{quote}
\small\ttfamily
\{ \\
\quad "keywords": ["keyword1", "keyword2"], \\
\quad "persons": ["name1", "name2"], \\
\quad "time\_expression": "time expression or null", \\
\quad "time\_range\_start": "YYYY-MM-DD or null", \\
\quad "time\_range\_end": "YYYY-MM-DD or null", \\
\quad "location": "location or null", \\
\quad "entities": ["entity1"] \\
\}
\end{quote}

\textbf{Constraint.}
Return only JSON.
\end{tcolorbox}

\begin{tcolorbox}[
    title={Retrieval Planning Prompt},
    colback=gray!3,
    colframe=gray!60,
    fonttitle=\bfseries,
    breakable
]
\small
\textbf{System.}
You are an intelligent information requirement analyst. You must output valid JSON format.

\vspace{0.4em}
\textbf{User.}
Analyze the following question and determine what information is required to answer it.

\vspace{0.4em}
\textbf{Question.}
\texttt{\{query\}}

\vspace{0.4em}
\textbf{Consider.}
\begin{enumerate}[leftmargin=1.5em,itemsep=0.15em]
    \item What type of question is this?
    \item What entities, events, or concepts need to be identified?
    \item What relationships need to be established?
    \item What minimal information would be sufficient to answer the question?
\end{enumerate}

\vspace{0.4em}
\textbf{Output Format.}
\begin{quote}
\small\ttfamily
\{ \\
\quad "question\_type": "factual / temporal / relational / explanatory / other", \\
\quad "key\_entities": ["entity1", "entity2"], \\
\quad "required\_info": [ \\
\quad\quad \{ \\
\quad\quad\quad "info\_type": "type of information", \\
\quad\quad\quad "description": "specific information needed", \\
\quad\quad\quad "priority": "high / medium / low" \\
\quad\quad \} \\
\quad ], \\
\quad "relationships": ["relationship1"], \\
\quad "minimal\_queries\_needed": 2 \\
\}
\end{quote}

\textbf{Constraint.}
Focus on the minimal essential information, not exhaustive details. Return only JSON.
\end{tcolorbox}

\begin{tcolorbox}[
    title={Answer Generation Prompt},
    colback=gray!3,
    colframe=gray!60,
    fonttitle=\bfseries,
    breakable
]
\small
\textbf{System.}
You are a professional Q\&A assistant. Extract concise answers from context. You must output valid JSON format.

\vspace{0.4em}
\textbf{Context Format.}
\begin{quote}
\small\ttfamily
[Context \{i\}] \\
Content: \{entry.lossless\_restatement\} \\
Session: \{entry.session\_date\} -> \{entry.session\_end\_date\} \\
Time: \{entry.timestamp\} \\
Location: \{entry.location\} \\
Persons: \{entry.persons\} \\
Related Entities: \{entry.entities\} \\
Topic: \{entry.topic\}
\end{quote}

\vspace{0.4em}
\textbf{User.}
Answer the user's question based on the provided context.

\vspace{0.4em}
\textbf{User Question.}
\texttt{\{query\}}

\vspace{0.4em}
\textbf{Relevant Context.}
\texttt{\{context\_str\}}

\vspace{0.4em}
\textbf{Requirements.}
\begin{enumerate}[leftmargin=1.5em,itemsep=0.15em]
    \item Think through the reasoning process.
    \item Provide a very concise answer.
    \item Answer only from the provided context.
    \item Format dates as ``DD Month YYYY'' when dates are needed.
    \item Return valid JSON.
\end{enumerate}

\vspace{0.4em}
\textbf{Output Format.}
\begin{quote}
\small\ttfamily
\{ \\
\quad "reasoning": "Brief explanation of the reasoning.", \\
\quad "answer": "Concise answer in a short phrase." \\
\}
\end{quote}

\textbf{Constraint.}
Return only JSON.
\end{tcolorbox}

\subsection{Analysis of Main Results}
\label{app:main_results_analysis}

Table~\ref{tab:locomo_full_results} presents the full LoCoMo results across four backbone models and four question categories. RaMem achieves the strongest average F1 on every backbone. On closed-source models, RaMem improves average F1 from 39.06 to 51.66 on GPT-4o and from 43.24 to 54.23 on GPT-4.1-mini. The improvement is also consistent on open-source models, where average F1 increases from 33.45 to 44.55 on Qwen3-8B and from 17.98 to 24.65 on Qwen2.5-3B. Average BLEU follows the same overall trend across all four backbones. These gains are meaningful because SimpleMem is already a strong structured-memory baseline that uses semantic compression, multi-view indexing, and intent-aware retrieval. RaMem therefore provides a complementary benefit: it improves the contextual validity of the retrieved evidence, rather than only making memory more compact or retrievable.

The gains are especially clear on stronger backbones. On GPT-4o, RaMem improves average F1 by 12.60 points and average BLEU by 16.35 points over SimpleMem. It also achieves the best F1 in MultiHop, Temporal, and OpenDomain categories, while remaining competitive with the full-history LoCoMo setting on SingleHop questions. On GPT-4.1-mini, RaMem improves average F1 by 10.99 points and average BLEU by 8.24 points. The largest category-level improvements appear in OpenDomain and SingleHop questions, where F1 increases from 19.76 to 26.35 and from 51.12 to 58.14, respectively. This suggests that contextual reinstatement is not merely helping temporally phrased questions. It also helps the model select the correct episode or state when multiple memories share entities or topics.

Open-source models show the same pattern. On Qwen3-8B, RaMem improves average F1 by 11.10 points and average BLEU by 8.33 points over SimpleMem. The method improves MultiHop F1 from 28.97 to 36.10 and SingleHop F1 from 46.62 to 50.59, showing that contextually valid retrieval benefits both compositional reasoning and direct factual lookup. On Qwen2.5-3B, RaMem improves average F1 from 17.98 to 24.65 and average BLEU from 14.89 to 19.44. Although this smaller model remains less capable overall, the consistent gain indicates that improving the evidence before generation is useful even when the generator itself is weak.

The category-level results also clarify the role of temporal and episodic information. RaMem does improve the Temporal category on most backbones, but its advantage is not confined to temporal reasoning. On GPT-4o, RaMem improves MultiHop, OpenDomain, and SingleHop F1 by 11.42, 7.35, and 7.46 points, respectively. On GPT-4.1-mini, the largest gains are in OpenDomain and SingleHop questions. On Qwen3-8B and Qwen2.5-3B, RaMem again improves MultiHop and SingleHop performance by clear margins. These results support the main design motivation: temporal and session information is not used only to answer temporal questions. It helps identify which episode, state, or event instance should be used as evidence across many types of memory queries.

There are also a few informative exceptions. For example, GPT-4.1-mini and Qwen3-8B show lower Temporal BLEU than SimpleMem, even though Temporal F1 improves. On Qwen2.5-3B, RaMem improves Temporal F1 over SimpleMem but has lower Temporal BLEU than SimpleMem and A-Mem. This suggests that once the correct evidence is retrieved, answer realization can still vary in wording or granularity, especially for smaller generators and temporal answers. The overall pattern nevertheless remains consistent: RaMem improves average answer quality and achieves broad gains across categories, indicating that contextual verifiability is a robust complement to existing memory compression and retrieval strategies.

Table~\ref{tab:longmemeval_results} further evaluates RaMem on LongMemEval-S, which stresses long-horizon memory under fine-grained temporal, update, and session-specific queries. RaMem achieves the best average accuracy, improving over SimpleMem from 76.87 to 80.15. The gains are strongest in categories that require identifying the correct memory episode or state, including Multi-Session, Single-Session-Assistant, and Single-Session-Preference. RaMem also slightly improves Temporal accuracy over the strongest prior result, while remaining competitive in Single-Session-User and Knowledge-Update categories where LightMem is especially strong. These results complement the LoCoMo findings: contextual reinstatement is useful not only for conversational QA with F1 or BLEU evaluation, but also under judge-based accuracy evaluation on a separate long-memory benchmark.

\begin{table}[t]
\centering
\scriptsize
\setlength{\tabcolsep}{4.5pt}
\renewcommand{\arraystretch}{1.08}
\caption{Performance comparison on LongMemEval-S. Following the SimpleMem protocol, we use \texttt{gpt-4.1-mini} as the LLM judge and report accuracy across temporal, multi-session, knowledge-update, and single-session memory categories.}
\label{tab:longmemeval_results}
\begin{tabular}{lccccc}
\toprule
\textbf{Category} & \textbf{Full-context} & \textbf{Mem0} & \textbf{LightMem} & \textbf{SimpleMem} & \textbf{RaMem}\\
\midrule
Temporal & 27.06 & 40.60 & \underline{85.71} & 83.46 & \textbf{86.47} \\
Multi-Session & 30.08 & 50.37 & 47.37 & \underline{60.92} & \textbf{65.41} \\
Knowledge-Update & 41.03 & 69.23 & \textbf{92.30} & 79.48 & \underline{83.33} \\
Single-Session-User & 47.14 & \underline{87.14} & \textbf{88.57} & 85.71 & \underline{87.14} \\
Single-Session-Assistant & 32.14 & 48.21 & 21.43 & \underline{75.00} & \textbf{78.57} \\
Single-Session-Preference & 60.00 & 63.33 & \underline{76.67} & \underline{76.67} & \textbf{80.00} \\
\midrule
Average & 39.57 & 59.81 & 68.67 & \underline{76.87} & \textbf{80.15} \\
\bottomrule
\end{tabular}
\end{table}

\subsection{Context Collapse Analysis}
\label{app:context_collapse}

\begin{table}[t]
\centering
\scriptsize
\setlength{\tabcolsep}{5pt}
\caption{Context collapse analysis on the context-confusable subset. D@1 measures whether the top retrieved memory is a context distractor. RankGap is the rank of the ground-truth memory minus the rank of the best context distractor, so lower is better.}
\label{tab:context_collapse_main}
\begin{tabular}{llccccc}
\toprule
\textbf{Backbone} & \textbf{Method} 
& \textbf{D@1} $\downarrow$ 
& \textbf{AvgD@5} $\downarrow$ 
& \textbf{RankGap} $\downarrow$ 
& \textbf{GT R@10} $\uparrow$ 
& \textbf{F1} $\uparrow$ \\
\midrule
\multirow{2}{*}{GPT-4.1-mini}
& SimpleMem & 0.4366 & 3.4514 & 5.4135 & 0.6842 & 0.5107 \\
& Ours      & \textbf{0.3495} & \textbf{3.1336} & \textbf{2.1502} & \textbf{0.7510} & \textbf{0.5290} \\
\midrule
\multirow{2}{*}{Qwen3-8B}
& SimpleMem & 0.4588 & 3.3909 & 6.1638 & 0.6036 & 0.4117 \\
& Ours      & \textbf{0.3846} & \textbf{3.0589} & \textbf{2.4664} & \textbf{0.6826} & \textbf{0.4489} \\
\midrule
\multirow{2}{*}{Qwen2.5-3B}
& SimpleMem & 0.5414 & 3.6317 & 6.4248 & 0.5676 & 0.2207 \\
& Ours      & \textbf{0.4796} & \textbf{3.3554} & \textbf{3.5679} & \textbf{0.6246} & \textbf{0.2281} \\
\bottomrule
\end{tabular}
\end{table}

We provide the complete context collapse analysis in Table~\ref{tab:context_collapse_full}. The analysis is conducted under two settings. The first setting uses all questions with verified ground-truth memories. The second setting focuses on the context-confusable subset, where the SimpleMem baseline retrieves at least one strict context distractor in its top-5 results. A strict context distractor is defined as a retrieved memory that is not the ground-truth memory, is content-related to the query or ground-truth memory according to person, entity, topic, or keyword overlap, and is contextually incompatible according to temporal or session mismatch. This rule-based detector does not use embedding similarity in the reported results, which makes the analysis conservative and interpretable.

Across all verified questions, contextual reinstatement consistently reduces distractor exposure while improving ground-truth retrieval. For GPT-4.1-mini, D@1 decreases from 0.4273 to 0.3428, AvgDTop5 decreases from 3.3785 to 3.0773, and RankGap decreases from 5.1934 to 2.0278. At the same time, GT R@10 increases from 0.6896 to 0.7550 and MRR increases from 0.4751 to 0.5323. The same pattern holds for Qwen3-8B, where D@1 decreases from 0.4416 to 0.3702, RankGap decreases from 5.6976 to 2.1427, and GT R@10 increases from 0.6144 to 0.6925. Qwen2.5-3B follows the same trend, although the final F1 gain is smaller, suggesting that weaker generators remain limited in their ability to use improved evidence. Overall, these results show that contextual reinstatement changes the composition and ordering of retrieved evidence, not only the final generation behavior.

The context-confusable subset provides a more targeted test of the proposed failure mode. By construction, SimpleMem retrieves at least one strict context distractor in the top-5 results for every question in this subset, so D@5 is 1.0000 for the baseline. The key question is whether our method can reduce the dominance of these distractors and recover the correct episode more effectively. The answer is consistently positive. On GPT-4.1-mini, D@1 decreases from 0.4366 to 0.3495, RankGap decreases from 5.4135 to 2.1502, and GT R@10 increases from 0.6842 to 0.7510. On Qwen3-8B, D@1 decreases from 0.4588 to 0.3846 and RankGap decreases from 6.1638 to 2.4664, while GT R@10 increases from 0.6036 to 0.6826. On Qwen2.5-3B, D@1 decreases from 0.5414 to 0.4796 and RankGap decreases from 6.4248 to 3.5679. These results support the interpretation that contextual reinstatement is most beneficial when semantic similarity is actively misleading. It reduces the probability that a context-invalid distractor appears as the leading evidence and makes the ground-truth memory more competitive under a fixed retrieval budget.

\begin{table*}[t]
\centering
\small
\caption{Full context collapse analysis. The baseline is SimpleMem. D@K measures whether at least one context distractor appears in the top-$K$ retrieved memories. AvgDTop5 is the average number of context distractors in the top-5 memories. RankGap is the rank of the ground-truth memory minus the rank of the best context distractor. Lower values are better for D@K, AvgDTop5, and RankGap.}
\label{tab:context_collapse_full}
\resizebox{\textwidth}{!}{
\begin{tabular}{lllcccccccccc}
\toprule
\textbf{Setting} & \textbf{Backbone} & \textbf{Method} & \textbf{N} & \textbf{D@1} $\downarrow$ & \textbf{D@3} $\downarrow$ & \textbf{D@5} $\downarrow$ & \textbf{AvgDTop5} $\downarrow$ & \textbf{RankGap} $\downarrow$ & \textbf{GT R@5} $\uparrow$ & \textbf{GT R@10} $\uparrow$ & \textbf{MRR} $\uparrow$ & \textbf{F1} $\uparrow$ \\
\midrule
\multirow{6}{*}{All Verified GT}
& \multirow{2}{*}{Qwen3-8B}
& SimpleMem & 1499 & 0.4416 & 0.8599 & 0.9626 & 3.2642 & 5.6976 & 0.5243 & 0.6144 & 0.3991 & 0.4138 \\
& & Ours      & 1499 & \textbf{0.3702} & \textbf{0.8025} & \textbf{0.9320} & \textbf{2.9533} & \textbf{2.1427} & \textbf{0.6011} & \textbf{0.6925} & \textbf{0.4502} & \textbf{0.4522} \\
\cmidrule(lr){2-13}
& \multirow{2}{*}{Qwen2.5-3B}
& SimpleMem & 1503 & 0.5223 & 0.8862 & 0.9647 & 3.5037 & 6.0253 & 0.4970 & 0.5768 & 0.3773 & 0.2250 \\
& & Ours      & 1502 & \textbf{0.4627} & \textbf{0.8329} & \textbf{0.9427} & \textbf{3.2550} & \textbf{3.3185} & \textbf{0.5473} & \textbf{0.6332} & \textbf{0.4112} & \textbf{0.2315} \\
\cmidrule(lr){2-13}
& \multirow{2}{*}{GPT-4.1-mini}
& SimpleMem & 1514 & 0.4273 & 0.8943 & 0.9789 & 3.3785 & 5.1934 & 0.6156 & 0.6896 & 0.4751 & 0.5109 \\
& & Ours      & 1514 & \textbf{0.3428} & \textbf{0.8309} & \textbf{0.9551} & \textbf{3.0773} & \textbf{2.0278} & \textbf{0.6737} & \textbf{0.7550} & \textbf{0.5323} & \textbf{0.5291} \\
\midrule
\multirow{6}{*}{Context-Confusable}
& \multirow{2}{*}{Qwen3-8B}
& SimpleMem & 1443 & 0.4588 & 0.8933 & 1.0000 & 3.3909 & 6.1638 & 0.5100 & 0.6036 & 0.3926 & 0.4117 \\
& & Ours      & 1443 & \textbf{0.3846} & \textbf{0.8337} & \textbf{0.9605} & \textbf{3.0589} & \textbf{2.4664} & \textbf{0.5925} & \textbf{0.6826} & \textbf{0.4452} & \textbf{0.4489} \\
\cmidrule(lr){2-13}
& \multirow{2}{*}{Qwen2.5-3B}
& SimpleMem & 1450 & 0.5414 & 0.9186 & 1.0000 & 3.6317 & 6.4248 & 0.4862 & 0.5676 & 0.3735 & 0.2207 \\
& & Ours      & 1449 & \textbf{0.4796} & \textbf{0.8585} & \textbf{0.9655} & \textbf{3.3554} & \textbf{3.5679} & \textbf{0.5411} & \textbf{0.6246} & \textbf{0.4094} & \textbf{0.2281} \\
\cmidrule(lr){2-13}
& \multirow{2}{*}{GPT-4.1-mini}
& SimpleMem & 1482 & 0.4366 & 0.9136 & 1.0000 & 3.4514 & 5.4135 & 0.6093 & 0.6842 & 0.4740 & 0.5107 \\
& & Ours      & 1482 & \textbf{0.3495} & \textbf{0.8468} & \textbf{0.9669} & \textbf{3.1336} & \textbf{2.1502} & \textbf{0.6694} & \textbf{0.7510} & \textbf{0.5330} & \textbf{0.5290} \\
\bottomrule
\end{tabular}
}
\end{table*}

\subsection{Context Shuffle Results}
\label{app:context_shuffle}

We provide complete context shuffle results in Tables~\ref{tab:context_shuffle_all_verified} and~\ref{tab:context_shuffle_temporal}. The goal of this experiment is to test whether the proposed method depends on a meaningful binding between memory content and episodic context. All shuffle variants keep the memory content unchanged, including the lossless restatement and its dense embedding. Only the context fields are shuffled across memory entries within the same memory bank. Thus, the corrupted variants preserve the same amount of metadata while breaking the correspondence between each memory and its original episodic coordinates.

\begin{table*}[t]
\centering
\small
\caption{Context shuffle test on all verified ground-truth questions. Lower is better for CappedRank@10, D@1, AvgDTop5, and RankGap.}
\label{tab:context_shuffle_all_verified}
\resizebox{\textwidth}{!}{
\begin{tabular}{llcccccccccc}
\toprule
\textbf{Backbone} & \textbf{Variant} 
& \textbf{N} 
& \textbf{F1} $\uparrow$ 
& \textbf{BLEU1} $\uparrow$ 
& \textbf{GT R@5} $\uparrow$ 
& \textbf{GT R@10} $\uparrow$ 
& \textbf{MRR} $\uparrow$ 
& \textbf{CappedRank@10} $\downarrow$ 
& \textbf{D@1} $\downarrow$ 
& \textbf{AvgDTop5} $\downarrow$ 
& \textbf{RankGap} $\downarrow$ \\
\midrule

\multirow{5}{*}{GPT-4.1-mini}
& Original & 1514 & 0.5291 & 0.4475 & 0.6737 & 0.7550 & 0.5323 & 5.5132 & 0.3428 & 3.0773 & 2.0278 \\
& Shuffle temporal & 1514 & 0.5004 & 0.4230 & 0.6103 & 0.6744 & 0.4871 & 7.6149 & 0.5753 & 4.0396 & 5.2974 \\
& Shuffle entity & 1514 & 0.5256 & 0.4444 & 0.6830 & 0.7523 & 0.5349 & 5.4465 & 0.3382 & 3.0918 & 1.9761 \\
& Shuffle topic & 1514 & 0.5255 & 0.4441 & 0.6777 & 0.7523 & 0.5327 & 5.4815 & 0.3408 & 3.0746 & 1.9861 \\
& Shuffle full context & 1514 & 0.5079 & 0.4296 & 0.6050 & 0.6638 & 0.4822 & 7.6876 & 0.5872 & 4.0707 & 5.3965 \\

\midrule

\multirow{5}{*}{Qwen3-8B}
& Original & 1499 & 0.4522 & 0.3801 & 0.6011 & 0.6925 & 0.4502 & 6.2028 & 0.3702 & 2.9533 & 2.1427 \\
& Shuffle temporal & 1499 & 0.4287 & 0.3606 & 0.5484 & 0.6204 & 0.4116 & 8.1134 & 0.6518 & 4.0747 & 5.7768 \\
& Shuffle entity & 1499 & 0.4526 & 0.3801 & 0.5997 & 0.6898 & 0.4517 & 6.2141 & 0.3689 & 3.0013 & 2.1699 \\
& Shuffle topic & 1499 & 0.4546 & 0.3828 & 0.6051 & 0.6898 & 0.4519 & 6.1654 & 0.3722 & 2.9533 & 2.0862 \\
& Shuffle full context & 1499 & 0.4347 & 0.3642 & 0.5437 & 0.6157 & 0.4053 & 8.1161 & 0.6631 & 4.1361 & 5.7634 \\
\bottomrule
\end{tabular}
}
\end{table*}

\begin{table*}[t]
\centering
\small
\caption{Context shuffle test on temporal-triggered questions. This subset isolates questions where temporal/session context is explicitly activated. Lower is better for CappedRank@10, D@1, AvgDTop5, and RankGap.}
\label{tab:context_shuffle_temporal}
\resizebox{\textwidth}{!}{
\begin{tabular}{llcccccccccc}
\toprule
\textbf{Backbone} & \textbf{Variant} 
& \textbf{N} 
& \textbf{F1} $\uparrow$ 
& \textbf{BLEU1} $\uparrow$ 
& \textbf{GT R@5} $\uparrow$ 
& \textbf{GT R@10} $\uparrow$ 
& \textbf{MRR} $\uparrow$ 
& \textbf{CappedRank@10} $\downarrow$ 
& \textbf{D@1} $\downarrow$ 
& \textbf{AvgDTop5} $\downarrow$ 
& \textbf{RankGap} $\downarrow$ \\
\midrule

\multirow{5}{*}{GPT-4.1-mini}
& Original & 212 & 0.6009 & 0.5396 & 0.6934 & 0.8066 & 0.5249 & 5.3113 & 0.2642 & 2.1462 & 1.0404 \\
& Shuffle temporal & 212 & 0.4985 & 0.4396 & 0.1981 & 0.2358 & 0.1796 & 21.0802 & 0.8160 & 4.2972 & 21.5029 \\
& Shuffle entity & 212 & 0.5843 & 0.5198 & 0.6887 & 0.7972 & 0.5184 & 5.4198 & 0.2830 & 2.2453 & 1.3266 \\
& Shuffle topic & 212 & 0.5982 & 0.5309 & 0.6887 & 0.8019 & 0.5171 & 5.4057 & 0.2736 & 2.1462 & 1.2211 \\
& Shuffle full context & 212 & 0.5180 & 0.4578 & 0.1887 & 0.2358 & 0.1762 & 21.3915 & 0.8302 & 4.3915 & 22.0116 \\

\midrule

\multirow{5}{*}{Qwen3-8B}
& Original & 212 & 0.5317 & 0.4810 & 0.5708 & 0.7170 & 0.4271 & 6.6038 & 0.2736 & 2.0142 & 1.2350 \\
& Shuffle temporal & 212 & 0.4081 & 0.3681 & 0.1651 & 0.1840 & 0.1250 & 20.3255 & 0.8726 & 4.3443 & 22.9595 \\
& Shuffle entity & 212 & 0.5075 & 0.4596 & 0.5613 & 0.7170 & 0.4238 & 6.5142 & 0.2736 & 2.0802 & 1.1585 \\
& Shuffle topic & 212 & 0.5195 & 0.4677 & 0.5660 & 0.7123 & 0.4286 & 6.4528 & 0.2783 & 2.0000 & 1.0546 \\
& Shuffle full context & 212 & 0.4134 & 0.3694 & 0.1557 & 0.1840 & 0.1242 & 20.3491 & 0.8868 & 4.5094 & 23.0952 \\
\bottomrule
\end{tabular}
}
\end{table*}

The all-question results show that corrupting temporal/session fields harms both retrieval and answer quality across different backbones. On GPT-4.1-mini, temporal shuffling reduces GT R@10 from 0.7550 to 0.6744 and increases D@1 from 0.3428 to 0.5753. RankGap also increases from 2.0278 to 5.2974, meaning that context-invalid distractors again dominate the ground-truth memory in the ranking. Qwen3-8B shows the same trend: GT R@10 drops from 0.6925 to 0.6204, D@1 increases from 0.3702 to 0.6518, and RankGap rises from 2.1427 to 5.7768. Full-context shuffling produces similar degradation on both backbones. These results indicate that correct temporal/session binding contributes to retrieval behavior across the benchmark, not only in a narrow subset of examples.

The temporal-triggered subset provides a sharper mechanism-level view. On GPT-4.1-mini, shuffling temporal fields causes GT R@10 to collapse from 0.8066 to 0.2358 and MRR from 0.5249 to 0.1796. At the same time, D@1 increases from 0.2642 to 0.8160 and RankGap rises from 1.0404 to 21.5029. Qwen3-8B exhibits the same behavior: GT R@10 drops from 0.7170 to 0.1840, MRR from 0.4271 to 0.1250, D@1 increases from 0.2736 to 0.8726, and RankGap rises from 1.2350 to 22.9595. Full-context shuffling closely matches this failure pattern, which suggests that the main source of degradation is the corrupted temporal/session coordinate. By contrast, shuffling entity or topic fields has much smaller effects. This finding clarifies the current implementation: contextual reinstatement is primarily operationalized through session-level temporal grounding, while entity and topic fields serve as auxiliary cues. The results show that structured context is useful only when it remains correctly bound to memory content, and that breaking this binding restores the context collapse that the method is designed to mitigate.

\subsection{Temporal Buffer Sensitivity}
\label{app:buffer_sensitivity}

Several consistent trends emerge. First, the best performance is obtained with a \emph{moderate} temporal buffer rather than with the strictest or widest setting. On GPT-4.1-mini, the 5-day window gives the strongest overall performance on all verified questions, achieving an F1 of 0.5291 and a GT R@10 of 0.7550, while 3--5 days form a similarly strong region on the temporal-triggered subset. On Qwen3-8B, the optimum shifts slightly toward a narrower setting, with 3 days producing the strongest temporal-triggered performance and near-best overall performance. This pattern is informative: the system benefits from some tolerance to event-time and mention-time mismatch, but the optimal tolerance is not unbounded. The results therefore support the design intuition of contextual reinstatement. Temporal context should function as a calibrated episodic coordinate, not as a hard equality constraint.

Second, enlarging the buffer too aggressively weakens contextual specificity and reintroduces distractors. This is most visible at 14 days. On GPT-4.1-mini, all-verified GT R@10 drops from 0.7550 at 5 days to 0.7417 at 14 days, while on the temporal-triggered subset RankGap increases from 1.0404 to 2.5736 and AvgDTop5 rises from 2.1462 to 2.3821. Qwen3-8B exhibits an even clearer degradation: on temporal-triggered questions, F1 drops from 0.5329 at 3 days to 0.4799 at 14 days, while GT R@10 decreases from 0.7311 to 0.6651. These changes show that a broader temporal window does not monotonically improve recall. Instead, it eventually admits more context-invalid memories and reduces the ranking advantage of the correct episode. Taken together, the two backbones tell a consistent story: contextual reinstatement requires a \emph{moderate} temporal window that absorbs small temporal mismatches while preserving contextual validity.

We provide the full temporal buffer sensitivity results in Tables~\ref{tab:buffer_all_verified} and~\ref{tab:buffer_temporal_triggered}. The goal of this experiment is to understand how the temporal reinstatement window affects the balance between recall and contextual specificity. A zero-day window is maximally strict and only admits temporally aligned memories, while larger windows allow increasing tolerance for mismatch between event time and mention time. In principle, a larger buffer can help recover memories that belong to the correct episode but were mentioned slightly outside the exact query span. However, an excessively broad buffer may also admit nearby but context-invalid memories, thereby increasing distractor exposure. We therefore evaluate buffer sizes of 0, 1, 3, 5, 7, and 14 days on both GPT-4.1-mini and Qwen3-8B.

\begin{table*}[t]
\centering
\small
\caption{Temporal buffer sensitivity on all verified questions. Lower is better for D@1, RankGap, and AvgDTop5.}
\label{tab:buffer_all_verified}
\resizebox{\textwidth}{!}{
\begin{tabular}{llccccccc}
\toprule
\textbf{Backbone} & \textbf{Buffer} & \textbf{N} & \textbf{F1} $\uparrow$ & \textbf{GT R@10} $\uparrow$ & \textbf{MRR} $\uparrow$ & \textbf{D@1} $\downarrow$ & \textbf{RankGap} $\downarrow$ & \textbf{AvgDTop5} $\downarrow$ \\
\midrule
\multirow{6}{*}{GPT-4.1-mini}
& 0  & 1513 & 0.5226 & 0.7396 & 0.5279 & 0.3410 & 1.6488 & 3.0330 \\
& 1  & 1514 & 0.5172 & 0.7490 & 0.5304 & 0.3461 & 1.5294 & 3.0297 \\
& 3  & 1514 & 0.5270 & 0.7483 & 0.5321 & 0.3435 & 1.7790 & 3.0396 \\
& 5  & 1514 & \textbf{0.5291} & \textbf{0.7550} & \textbf{0.5323} & 0.3428 & 2.0278 & 3.0773 \\
& 7  & 1514 & 0.5245 & 0.7503 & 0.5340 & \textbf{0.3415} & 1.9992 & 3.0727 \\
& 14 & 1514 & 0.5224 & 0.7417 & 0.5302 & 0.3428 & 2.1756 & 3.0991 \\
\midrule
\multirow{6}{*}{Qwen3-8B}
& 0  & 1499 & 0.4496 & 0.6905 & 0.4521 & 0.3682 & 1.6099 & 2.9333 \\
& 1  & 1499 & \textbf{0.4558} & 0.6918 & 0.4523 & \textbf{0.3649} & 1.5397 & \textbf{2.9260} \\
& 3  & 1499 & 0.4557 & \textbf{0.6938} & \textbf{0.4528} & \textbf{0.3649} & 1.7316 & 2.9386 \\
& 5  & 1499 & 0.4522 & 0.6925 & 0.4502 & 0.3702 & 2.1427 & 2.9533 \\
& 7  & 1499 & 0.4532 & 0.6865 & 0.4465 & 0.3696 & 2.2101 & 2.9606 \\
& 14 & 1499 & 0.4499 & 0.6865 & 0.4485 & 0.3769 & 2.2721 & 2.9960 \\
\bottomrule
\end{tabular}
}
\end{table*}

\begin{table*}[t]
\centering
\small
\caption{Temporal buffer sensitivity on the temporal-triggered subset. Lower is better for D@1, RankGap, and AvgDTop5.}
\label{tab:buffer_temporal_triggered}
\resizebox{\textwidth}{!}{
\begin{tabular}{llccccccc}
\toprule
\textbf{Backbone} & \textbf{Buffer} & \textbf{N} & \textbf{F1} $\uparrow$ & \textbf{GT R@10} $\uparrow$ & \textbf{MRR} $\uparrow$ & \textbf{D@1} $\downarrow$ & \textbf{RankGap} $\downarrow$ & \textbf{AvgDTop5} $\downarrow$ \\
\midrule
\multirow{6}{*}{GPT-4.1-mini}
& 0  & 212 & 0.5954 & 0.7358 & 0.4837 & 0.2783 & -0.5663 & 1.9481 \\
& 1  & 212 & 0.5841 & 0.7877 & 0.5196 & \textbf{0.2594} & -1.9137 & \textbf{1.9198} \\
& 3  & 212 & \textbf{0.6146} & 0.7925 & 0.5126 & 0.2689 & -0.2929 & 2.0236 \\
& 5  & 212 & 0.6009 & \textbf{0.8066} & 0.5249 & 0.2642 & 1.0404 & 2.1462 \\
& 7  & 212 & 0.5931 & 0.7925 & \textbf{0.5250} & 0.2830 & 1.7828 & 2.2123 \\
& 14 & 212 & 0.5906 & 0.7547 & 0.5088 & 0.2925 & 2.5736 & 2.3821 \\
\midrule
\multirow{6}{*}{Qwen3-8B}
& 0  & 212 & 0.4987 & 0.6934 & 0.4145 & 0.2736 & -1.5543 & 1.8679 \\
& 1  & 212 & 0.5187 & 0.7217 & 0.4293 & 0.2594 & -2.3118 & \textbf{1.8019} \\
& 3  & 212 & \textbf{0.5329} & \textbf{0.7311} & \textbf{0.4332} & \textbf{0.2547} & -0.6237 & 1.8632 \\
& 5  & 212 & 0.5317 & 0.7170 & 0.4271 & 0.2736 & 1.2350 & 2.0142 \\
& 7  & 212 & 0.5231 & 0.7028 & 0.4235 & 0.2689 & 1.7611 & 2.0708 \\
& 14 & 212 & 0.4799 & 0.6651 & 0.4066 & 0.3066 & 2.9257 & 2.3443 \\
\bottomrule
\end{tabular}
}
\end{table*}

The results show that the temporal reinstatement window should be neither maximally strict nor excessively broad. On all verified questions, both backbones achieve their strongest or near-strongest performance in the moderate range between 3 and 5 days. GPT-4.1-mini performs best at 5 days, where F1 reaches 0.5291 and GT R@10 reaches 0.7550. Qwen3-8B is slightly more conservative, with 1--3 days giving the strongest overall results and 3 days producing the best GT R@10. This pattern is consistent with the intended role of the buffer. Some tolerance is needed to absorb mild mismatch between event time and mention time, but once the window becomes too large, the system begins to retrieve less specific evidence.

The trade-off is much clearer on the temporal-triggered subset, where temporal reinstatement is actually used to recover an episode. Here, moderate windows again dominate. GPT-4.1-mini achieves its highest F1 at 3 days and its highest GT R@10 at 5 days, while Qwen3-8B peaks at 3 days on F1, GT R@10, MRR, and D@1. In contrast, the 14-day setting consistently degrades performance. For GPT-4.1-mini, temporal-triggered RankGap increases from 1.0404 at 5 days to 2.5736 at 14 days, and AvgDTop5 increases from 2.1462 to 2.3821. For Qwen3-8B, F1 drops from 0.5329 at 3 days to 0.4799 at 14 days, while GT R@10 decreases from 0.7311 to 0.6651 and D@1 rises from 0.2547 to 0.3066. These results support a clear interpretation: a moderate temporal window offers the best balance between tolerating temporal mismatch and avoiding context-invalid distractors. In other words, contextual reinstatement benefits from calibrated flexibility, not from indiscriminately widening the search span.

\subsection{Retrieval Budget Sensitivity}
\label{app:budget_sensitivity}

\begin{figure*}[t]
    \centering
    \includegraphics[width=\textwidth]{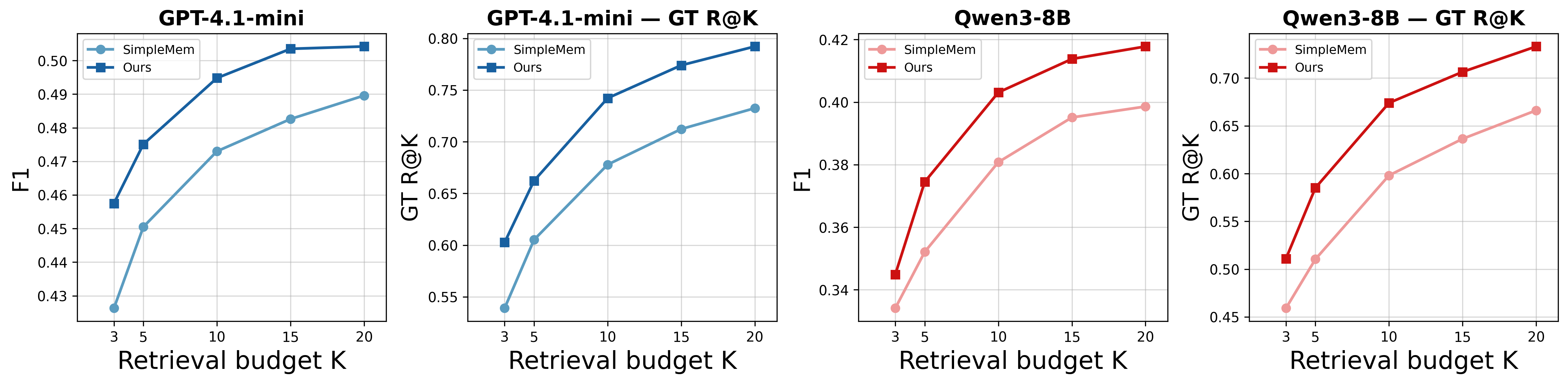}
    \caption{Sensitivity to retrieval budget. Each subplot varies the number of retrieved memories passed to the generator and compares SimpleMem with our method. Across both backbones, contextual reinstatement consistently improves F1 and GT R@$K$ under all budgets, with particularly clear advantages in the low-budget regime.}
    \label{fig:budget_sensitivity}
\end{figure*}

To evaluate whether contextual reinstatement remains effective under limited context budgets, we vary the number of retrieved memories passed to the generator and compare our method with the SimpleMem baseline. This experiment is important because the main claim of our framework is not merely that it retrieves more memories, but that it ranks \emph{better} memories earlier. If this claim is correct, our advantage should already appear under small budgets, where the generator can only observe a few retrieved entries. Figure~\ref{fig:budget_sensitivity} reports the trends on GPT-4.1-mini and Qwen3-8B, using F1 and budget-aligned ground-truth recall (GT R@$K$) to capture both final answer quality and retrieval effectiveness. Two clear patterns emerge. First, contextual reinstatement consistently outperforms the baseline across the entire budget range. On GPT-4.1-mini, our method improves F1 from 0.4575 to 0.5042 as $K$ increases from 3 to 20, while remaining above the baseline at every budget. The same trend holds for budget-aligned ground-truth recall, where our method improves from 0.6026 to 0.7922 compared with the baseline range of 0.5390 to 0.7325. Qwen3-8B exhibits the same behavior: our method dominates the baseline on both F1 and GT R@$K$ for every retrieval budget. This result supports the main intuition of contextual reinstatement. The gain does not depend on retrieving a large amount of evidence. Instead, it appears already at small budgets, showing that the method improves the ordering of retrieved memories and makes the answer-supporting episode more accessible early in the ranked list.

Second, the low-budget regime is especially informative. Under tight budgets, the generator must rely on only a few retrieved memories, so ranking quality matters most. Here the advantage of our method is particularly meaningful. On GPT-4.1-mini, our method at $K=10$ already exceeds the baseline at $K=20$ on both F1 (0.4948 vs.\ 0.4896) and GT R@$K$ (0.7422 vs.\ 0.7325). The same pattern appears on Qwen3-8B, where our method at $K=10$ surpasses the baseline at $K=20$ on both F1 (0.4031 vs.\ 0.3986) and GT R@$K$ (0.6740 vs.\ 0.6662). This shows that contextual reinstatement improves not only answer quality but also \emph{context efficiency}: it can reach the performance of a larger retrieval budget while using fewer retrieved memories. In practical agent systems, where context budget is often constrained, this behavior is especially valuable.

We provide the full retrieval-budget sensitivity results in Table~\ref{tab:budget_sensitivity_full}. The purpose of this experiment is to test whether contextual reinstatement improves only the total amount of retrieved evidence, or whether it also improves the ranking quality of memories under constrained budgets. We vary the number of retrieved memories passed to the generator from $K=3$ to $K=20$, and report final answer quality, budget-aligned ground-truth recall (GT R@$K$), MRR, and two context-collapse diagnostics. Because D@1 measures whether the top-ranked memory is a context distractor, it is unaffected by the truncation budget and remains constant for a fixed retrieval system. RankGap is also reported for completeness, although it is most informative when comparing methods at the same budget rather than when comparing different budgets.

\begin{table*}[t]
\centering
\small
\caption{Sensitivity to retrieval budget. GT R@$K$ denotes the fraction of questions whose ground-truth memory is included within the top-$K$ retrieved memories. Lower is better for D@1 and RankGap.}
\label{tab:budget_sensitivity_full}
\resizebox{\textwidth}{!}{
\begin{tabular}{llccccccc}
\toprule
\textbf{Backbone} & \textbf{Method / $K$} & \textbf{F1} $\uparrow$ & \textbf{BLEU1} $\uparrow$ & \textbf{GT R@$K$} $\uparrow$ & \textbf{MRR} $\uparrow$ & \textbf{D@1} $\downarrow$ & \textbf{RankGap} $\downarrow$ \\
\midrule
\multirow{10}{*}{GPT-4.1-mini}
& SimpleMem, $K=3$  & 0.4263 & 0.3530 & 0.5390 & 0.4344 & 0.4201 & -0.5637 \\
& SimpleMem, $K=5$  & 0.4505 & 0.3739 & 0.6052 & 0.4496 & 0.4201 & -0.4452 \\
& SimpleMem, $K=10$ & 0.4730 & 0.3942 & 0.6779 & 0.4593 & 0.4201 & 0.1772 \\
& SimpleMem, $K=15$ & 0.4826 & 0.4033 & 0.7123 & 0.4620 & 0.4201 & 0.7028 \\
& SimpleMem, $K=20$ & 0.4896 & 0.4107 & 0.7325 & 0.4632 & 0.4201 & 1.1392 \\
& Ours, $K=3$       & 0.4575 & 0.3793 & 0.6026 & 0.4932 & 0.3370 & -0.7108 \\
& Ours, $K=5$       & 0.4751 & 0.3985 & 0.6623 & 0.5069 & 0.3370 & -0.7696 \\
& Ours, $K=10$      & 0.4948 & 0.4167 & 0.7422 & 0.5180 & 0.3370 & -0.2264 \\
& Ours, $K=15$      & 0.5035 & 0.4244 & 0.7740 & 0.5205 & 0.3370 & 0.1835 \\
& Ours, $K=20$      & 0.5042 & 0.4228 & 0.7922 & 0.5215 & 0.3370 & 0.5321 \\
\midrule
\multirow{10}{*}{Qwen3-8B}
& SimpleMem, $K=3$  & 0.3341 & 0.2721 & 0.4591 & 0.3564 & 0.4299 & -0.4991 \\
& SimpleMem, $K=5$  & 0.3521 & 0.2881 & 0.5104 & 0.3679 & 0.4299 & -0.5054 \\
& SimpleMem, $K=10$ & 0.3808 & 0.3144 & 0.5981 & 0.3798 & 0.4299 & 0.2013 \\
& SimpleMem, $K=15$ & 0.3951 & 0.3269 & 0.6364 & 0.3828 & 0.4299 & 0.8306 \\
& SimpleMem, $K=20$ & 0.3986 & 0.3311 & 0.6662 & 0.3845 & 0.4299 & 1.5361 \\
& Ours, $K=3$       & 0.3449 & 0.2822 & 0.5110 & 0.4032 & 0.3604 & -0.6649 \\
& Ours, $K=5$       & 0.3745 & 0.3107 & 0.5851 & 0.4203 & 0.3604 & -0.6400 \\
& Ours, $K=10$      & 0.4031 & 0.3329 & 0.6740 & 0.4323 & 0.3604 & -0.1169 \\
& Ours, $K=15$      & 0.4138 & 0.3422 & 0.7065 & 0.4350 & 0.3604 & 0.2999 \\
& Ours, $K=20$      & 0.4178 & 0.3477 & 0.7331 & 0.4364 & 0.3604 & 0.8714 \\
\bottomrule
\end{tabular}
}
\end{table*}

The results show that contextual reinstatement improves performance under every tested retrieval budget. On GPT-4.1-mini, the baseline F1 rises from 0.4263 at $K=3$ to 0.4896 at $K=20$, while our method rises from 0.4575 to 0.5042. The same consistent dominance appears in GT R@$K$, where our method improves from 0.6026 to 0.7922 compared with the baseline range of 0.5390 to 0.7325. Qwen3-8B follows the same pattern: our method improves F1 from 0.3449 to 0.4178 and GT R@$K$ from 0.5110 to 0.7331, remaining above the baseline at every budget. MRR is also consistently higher for our method on both backbones, which indicates that the answer-supporting memory is ranked closer to the top of the retrieved list. D@1 remains lower for our method on both backbones, showing that the top retrieved evidence is less likely to be a context distractor. Together, these results confirm that the gains are not tied to a specific budget size.

More importantly, the experiment reveals a clear efficiency advantage. On GPT-4.1-mini, our method with $K=10$ already exceeds the baseline with $K=20$ on both F1 (0.4948 vs.\ 0.4896) and GT R@$K$ (0.7422 vs.\ 0.7325). The same effect holds on Qwen3-8B, where our method at $K=10$ surpasses the baseline at $K=20$ on F1 (0.4031 vs.\ 0.3986) and GT R@$K$ (0.6740 vs.\ 0.6662). This means that contextual reinstatement does more than improve final accuracy: it allows the system to reach the same or better performance using fewer retrieved memories. Such behavior is particularly valuable in practical long-context agent systems, where retrieval budget is tightly coupled to latency and prompt cost. The results therefore support a stronger claim than backbone-level improvement alone: contextual reinstatement yields a better performance--budget trade-off because it improves the ordering and contextual validity of the retrieved evidence.

\subsection{Efficiency and Cost Results}
\label{app:efficiency_cost}

\begin{table}[t]
\centering
\footnotesize
\setlength{\tabcolsep}{5pt}
\caption{Context-budget efficiency. RaMem with $K=10$ reaches or exceeds SimpleMem with $K=20$ while using about half the retrieved memory budget.}
\label{tab:budget_efficiency_main}
\begin{tabular}{llcccc}
\toprule
\textbf{Backbone} & \textbf{Method} & \textbf{K} 
& \textbf{F1} $\uparrow$ 
& \textbf{GT R@$K$} $\uparrow$ 
& \textbf{CtxTok/Q} $\downarrow$ \\
\midrule
\multirow{2}{*}{GPT-4.1-mini}
& SimpleMem & 20 & 0.4896 & 0.7325 & 1520.71 \\
& RaMem     & 10 & \textbf{0.4948} & \textbf{0.7422} & \textbf{758.24} \\
\midrule
\multirow{2}{*}{Qwen3-8B}
& SimpleMem & 20 & 0.3986 & 0.6662 & 1367.86 \\
& RaMem     & 10 & \textbf{0.4031} & \textbf{0.6740} & \textbf{683.50} \\
\bottomrule
\end{tabular}
\end{table}

We provide additional efficiency and cost results in Tables~\ref{tab:efficiency_default_full}, \ref{tab:budget_efficiency_full}, and~\ref{tab:timing_sample_full}. All runtime numbers compare SimpleMem-style retrieval and our method under the same optimized implementation. They should not be interpreted as a direct comparison against the original SimpleMem codebase, since our system uses a different serving stack with batching, parallel execution, and other engineering optimizations. The purpose of this analysis is to estimate the incremental cost of contextual reinstatement under a shared runtime environment.

\begin{table*}[t]
\centering
\small
\caption{Default online cost under the same optimized serving stack. Generation time and total time are unavailable in the historical full-evaluation logs and are therefore marked as NA.}
\label{tab:efficiency_default_full}
\resizebox{\textwidth}{!}{
\begin{tabular}{llcccccccc}
\toprule
\textbf{Backbone} & \textbf{Method}
& \textbf{F1} $\uparrow$
& \textbf{GT R@10} $\uparrow$
& \textbf{LLM Calls/Q} $\downarrow$
& \textbf{Retrieval Time/Q} $\downarrow$
& \textbf{Generation Time/Q} $\downarrow$
& \textbf{Total Time/Q} $\downarrow$
& \textbf{ContextChars/Q} $\downarrow$
& \textbf{Retrieved Memories/Q} $\downarrow$ \\
\midrule
\multirow{2}{*}{GPT-4.1-mini}
& SimpleMem & 0.5104 & 0.6779 & 4.0000 & 4.8346s & NA & NA & 14692.86 & 48.59 \\
& Ours      & \textbf{0.5281} & \textbf{0.7422} & 4.0000 & 4.9076s & NA & NA & \textbf{13261.07} & \textbf{43.86} \\
\midrule
\multirow{2}{*}{Qwen3-8B}
& SimpleMem & 0.4105 & 0.5981 & 4.0000 & 7.2318s & NA & NA & 12715.18 & 46.72 \\
& Ours      & \textbf{0.4459} & \textbf{0.6740} & 4.0000 & \textbf{6.6148s} & NA & NA & \textbf{11399.62} & \textbf{41.91} \\
\bottomrule
\end{tabular}
}
\end{table*}

\begin{table*}[t]
\centering
\small
\caption{Budget efficiency under different retrieval budgets. GT R@$K$ denotes the fraction of questions whose ground-truth memory is included within the top-$K$ retrieved memories.}
\label{tab:budget_efficiency_full}
\resizebox{\textwidth}{!}{
\begin{tabular}{lllccccc}
\toprule
\textbf{Backbone} & \textbf{Method} & \textbf{K}
& \textbf{F1} $\uparrow$
& \textbf{GT R@$K$} $\uparrow$
& \textbf{MRR} $\uparrow$
& \textbf{ContextTokens/Q} $\downarrow$
& \textbf{Retrieved Memories/Q} $\downarrow$ \\
\midrule
\multirow{5}{*}{GPT-4.1-mini}
& Ours      & 3  & 0.4575 & 0.6026 & 0.4932 & 226.48 & 3 \\
& Ours      & 5  & 0.4751 & 0.6623 & 0.5069 & 377.29 & 5 \\
& Ours      & 10 & 0.4948 & 0.7422 & 0.5180 & 758.24 & 10 \\
& SimpleMem & 10 & 0.4730 & 0.6779 & 0.4593 & 759.49 & 10 \\
& SimpleMem & 20 & 0.4896 & 0.7325 & 0.4632 & 1520.71 & 20 \\
\midrule
\multirow{3}{*}{Qwen3-8B}
& Ours      & 10 & 0.4031 & 0.6740 & 0.4323 & 683.50 & 10 \\
& SimpleMem & 10 & 0.3808 & 0.5981 & 0.3798 & 685.16 & 10 \\
& SimpleMem & 20 & 0.3986 & 0.6662 & 0.3845 & 1367.86 & 20 \\
\bottomrule
\end{tabular}
}
\end{table*}

\begin{table*}[t]
\centering
\small
\caption{Small-sample timing sanity check under the optimized serving stack. These results are included for transparency and should be interpreted as within-stack timing estimates rather than definitive throughput benchmarks.}
\label{tab:timing_sample_full}
\resizebox{\textwidth}{!}{
\begin{tabular}{llcccccccc}
\toprule
\textbf{Backbone} & \textbf{Method}
& \textbf{F1} $\uparrow$
& \textbf{GT R@10} $\uparrow$
& \textbf{LLM Calls/Q}
& \textbf{Retrieval Time/Q} $\downarrow$
& \textbf{Generation Time/Q} $\downarrow$
& \textbf{Total Time/Q} $\downarrow$
& \textbf{Retrieved Memories/Q} $\downarrow$
& \textbf{N} \\
\midrule
\multirow{2}{*}{GPT-4.1-mini}
& SimpleMem & 0.5104 & 0.6779 & 4.0000 & 5.3867s & 1.6344s & 7.0210s & 50.81 & 16 \\
& Ours      & \textbf{0.5281} & \textbf{0.7422} & 4.0000 & \textbf{5.2808s} & \textbf{1.5286s} & \textbf{6.8093s} & \textbf{44.25} & 16 \\
\midrule
\multirow{2}{*}{Qwen3-8B}
& SimpleMem & 0.4105 & 0.5981 & 4.0000 & 3.5873s & 1.3234s & 4.9107s & 51.88 & 16 \\
& Ours      & \textbf{0.4459} & \textbf{0.6740} & 4.0000 & \textbf{3.4441s} & \textbf{1.1860s} & \textbf{4.6302s} & \textbf{44.31} & 16 \\
\bottomrule
\end{tabular}
}
\end{table*}

The default-cost results show that contextual reinstatement does not require additional LLM calls under our planning-on and reflection-off setting. Both SimpleMem-style retrieval and our method use approximately four LLM calls per question. The added contextual operations are mainly structured filtering, temporal/session compatibility checking, and reranking, which are lightweight compared with LLM generation. On GPT-4.1-mini, retrieval time changes only marginally, while context length and retrieved memory count decrease. On Qwen3-8B, retrieval time also decreases under the same optimized stack. Since generation and total wall-clock time were not stored in the full historical logs, those columns are marked as NA in Table~\ref{tab:efficiency_default_full}.

The budget-efficiency results provide a more robust view of cost effectiveness. On both backbones, our method at $K=10$ reaches or exceeds SimpleMem at $K=20$ while using approximately half the retrieved memory budget. For GPT-4.1-mini, our method at $K=10$ achieves F1 of 0.4948 and GT R@$K$ of 0.7422, compared with 0.4896 and 0.7325 for SimpleMem at $K=20$. For Qwen3-8B, our method at $K=10$ achieves F1 of 0.4031 and GT R@$K$ of 0.6740, compared with 0.3986 and 0.6662 for SimpleMem at $K=20$. This indicates that contextual reinstatement improves the ranking quality and contextual validity of retrieved memories, allowing the generator to receive fewer but more useful memory entries.

Finally, the small-sample timing sanity check in Table~\ref{tab:timing_sample_full} is consistent with the default-cost analysis. Under the same optimized stack, our method does not increase total online time and often reduces it slightly because fewer memories are passed to the generator. However, these timing numbers are based on a small sample and should be interpreted as supporting evidence rather than as a standalone runtime benchmark. The main efficiency conclusion is therefore based on controlled within-stack cost comparison and context-budget efficiency, rather than direct runtime comparison to external codebases.

\subsection{Full Component Analysis}
\label{app:ablation}

We provide the complete ablation results in Tables~\ref{tab:ablation_gpt41mini} and~\ref{tab:ablation_qwen3_8b}. The main text reports the most diagnostic ablations on the temporal-triggered subset, while this appendix includes both all verified questions and temporal-triggered questions, as well as the cue-guard ablation. The cue guard disables a deterministic reliability check for contextual activation. In our experiments, this ablation has relatively small effects compared with removing session-level evidence conditions or context-aware ranking, so we treat it as a safety component rather than the primary source of the observed improvements.

\begin{table*}[t]
\centering
\small
\caption{Full ablation results on GPT-4.1-mini. Lower is better for D@1, AvgDTop5, and RankGap.}
\label{tab:ablation_gpt41mini}
\resizebox{\textwidth}{!}{
\begin{tabular}{llcccccccc}
\toprule
\textbf{Subset} & \textbf{Variant}
& \textbf{N}
& \textbf{F1} $\uparrow$
& \textbf{BLEU1} $\uparrow$
& \textbf{GT R@10} $\uparrow$
& \textbf{MRR} $\uparrow$
& \textbf{D@1} $\downarrow$
& \textbf{AvgDTop5} $\downarrow$
& \textbf{RankGap} $\downarrow$ \\
\midrule
\multirow{5}{*}{All Verified GT}
& Full & 1514 & \textbf{0.5298} & \textbf{0.4473} & \textbf{0.7483} & \textbf{0.5335} & \textbf{0.3435} & \textbf{3.0581} & \textbf{2.7454} \\
& w/o Session Context & 1514 & 0.5253 & 0.4442 & 0.6909 & 0.4857 & 0.4069 & 3.3963 & 5.5944 \\
& w/o Cue Guard & 1514 & 0.5289 & 0.4472 & 0.7351 & 0.5191 & 0.3567 & 3.0859 & 3.2846 \\
& w/o Context-Aware Ranking & 1514 & 0.5262 & 0.4459 & 0.6915 & 0.4781 & 0.4128 & 3.3943 & 5.9129 \\
& w/o Context-Preserved Generation & 1514 & 0.5133 & 0.4346 & \textbf{0.7483} & \textbf{0.5335} & \textbf{0.3435} & \textbf{3.0581} & \textbf{2.7454} \\
\midrule
\multirow{5}{*}{Temporal-triggered}
& Full & 212 & 0.5957 & \textbf{0.5319} & \textbf{0.8066} & 0.5205 & \textbf{0.2736} & 2.1509 & 1.0354 \\
& w/o Session Context & 212 & 0.5596 & 0.4922 & 0.5708 & 0.3512 & 0.5189 & 3.6132 & 10.0492 \\
& w/o Cue Guard & 212 & \textbf{0.5970} & \textbf{0.5319} & 0.8019 & 0.5203 & 0.2783 & \textbf{2.1462} & \textbf{0.9848} \\
& w/o Context-Aware Ranking & 212 & 0.5693 & 0.5018 & 0.5708 & 0.3468 & 0.5283 & 3.5896 & 12.6332 \\
& w/o Context-Preserved Generation & 212 & 0.5626 & 0.4968 & \textbf{0.8066} & 0.5205 & \textbf{0.2736} & 2.1509 & 1.0354 \\
\bottomrule
\end{tabular}
}
\end{table*}

\begin{table*}[t]
\centering
\small
\caption{Full ablation results on Qwen3-8B. Lower is better for D@1, AvgDTop5, and RankGap.}
\label{tab:ablation_qwen3_8b}
\resizebox{\textwidth}{!}{
\begin{tabular}{llcccccccc}
\toprule
\textbf{Subset} & \textbf{Variant}
& \textbf{N}
& \textbf{F1} $\uparrow$
& \textbf{BLEU1} $\uparrow$
& \textbf{GT R@10} $\uparrow$
& \textbf{MRR} $\uparrow$
& \textbf{D@1} $\downarrow$
& \textbf{AvgDTop5} $\downarrow$
& \textbf{RankGap} $\downarrow$ \\
\midrule
\multirow{5}{*}{All Verified GT}
& Full & 1499 & 0.4521 & 0.3807 & \textbf{0.6938} & \textbf{0.4519} & \textbf{0.3716} & \textbf{2.9460} & \textbf{2.0695} \\
& w/o Session Context & 1499 & 0.4423 & 0.3705 & 0.6071 & 0.4008 & 0.4396 & 3.2602 & 5.8922 \\
& w/o Cue Guard & 1499 & \textbf{0.4546} & \textbf{0.3831} & 0.6838 & 0.4467 & 0.3769 & 2.9740 & 2.6675 \\
& w/o Context-Aware Ranking & 1499 & 0.4448 & 0.3745 & 0.6177 & 0.4062 & 0.4336 & 3.2548 & 6.2023 \\
& w/o Context-Preserved Generation & 1499 & 0.4077 & 0.3391 & \textbf{0.6938} & \textbf{0.4519} & \textbf{0.3716} & \textbf{2.9460} & \textbf{2.0695} \\
\midrule
\multirow{5}{*}{Temporal-triggered}
& Full & 212 & 0.5226 & 0.4695 & \textbf{0.7217} & 0.4253 & 0.2783 & 2.0142 & 1.1685 \\
& w/o Session Context & 212 & 0.4578 & 0.4097 & 0.4245 & 0.2482 & 0.5849 & 3.4906 & 11.9565 \\
& w/o Cue Guard & 212 & \textbf{0.5292} & \textbf{0.4781} & 0.7170 & \textbf{0.4270} & \textbf{0.2736} & \textbf{2.0094} & \textbf{1.0489} \\
& w/o Context-Aware Ranking & 212 & 0.4642 & 0.4162 & 0.4434 & 0.2578 & 0.5755 & 3.5472 & 14.7735 \\
& w/o Context-Preserved Generation & 212 & 0.4248 & 0.3567 & \textbf{0.7217} & 0.4253 & 0.2783 & 2.0142 & 1.1685 \\
\bottomrule
\end{tabular}
}
\end{table*}

The complete results confirm the pattern discussed in the main text. Removing session context or context-aware ranking consistently harms retrieval quality. On all verified questions, these ablations reduce GT R@10 and MRR while increasing D@1, AvgDTop5, and RankGap. The effect is substantially stronger on temporal-triggered questions, where session-level evidence conditions are directly used to verify whether a memory belongs to the correct episode. For example, on Qwen3-8B, removing session context reduces GT R@10 from 0.7217 to 0.4245, while removing context-aware ranking reduces it to 0.4434. This shows that session grounding and validity-aware ranking are the two most important retrieval-side components.

The ablation of context-preserved generation behaves differently. It leaves retrieval metrics unchanged, because the retrieved memories are the same, but reduces final answer quality. This confirms that preserving episodic context is not only useful during retrieval; it also helps the generator interpret the selected evidence. The cue-guard ablation has smaller and less consistent effects. In some settings, final F1 is slightly higher without the guard, but retrieval diagnostics are generally similar or mildly worse. This suggests that the guard mainly serves as a lightweight reliability mechanism for avoiding false contextual constraints, while the main gains come from session-level evidence anchoring, validity-aware ranking, and context-preserved synthesis.

\subsection{Efficiency and Cost Analysis}
\label{app:efficiency_cost}

We further analyze whether contextual reinstatement improves memory quality at the cost of additional online overhead. Since our implementation uses an optimized serving stack with vLLM-style batching, parallel evaluation, and other engineering optimizations, we do not compare wall-clock runtime against the original SimpleMem codebase. Instead, we compare SimpleMem-style retrieval and our method under the same optimized implementation. This provides a fair estimate of the incremental cost introduced by contextual reinstatement. As shown in Table~\ref{tab:efficiency_main}, both methods use the same number of LLM calls per question under the planning-on and reflection-off setting. Contextual reinstatement does not introduce iterative reflection or extra generation calls. On GPT-4.1-mini, retrieval time changes only marginally from 4.83s to 4.91s per question, while F1 improves from 0.5104 to 0.5281 and GT R@10 improves from 0.6779 to 0.7422. On Qwen3-8B, our method is slightly faster in the same runtime stack, with retrieval time decreasing from 7.23s to 6.61s, while F1 improves from 0.4105 to 0.4459. The retrieved context is also smaller in both settings, reducing average context characters from 14.7K to 13.3K on GPT-4.1-mini and from 12.7K to 11.4K on Qwen3-8B.

\begin{table*}[h]
\centering
\small
\caption{Online efficiency under the same optimized serving stack. We compare SimpleMem-style retrieval and our method within the same implementation, rather than against the original SimpleMem runtime. Contextual reinstatement improves answer quality and retrieval quality without increasing LLM calls.}
\label{tab:efficiency_main}
\resizebox{\textwidth}{!}{
\begin{tabular}{llcccccc}
\toprule
\textbf{Backbone} & \textbf{Method} 
& \textbf{F1} $\uparrow$
& \textbf{GT R@10} $\uparrow$
& \textbf{LLM Calls/Q} $\downarrow$
& \textbf{Retrieval Time/Q} $\downarrow$
& \textbf{ContextChars/Q} $\downarrow$
& \textbf{Retrieved Memories/Q} $\downarrow$ \\
\midrule
\multirow{2}{*}{GPT-4.1-mini}
& SimpleMem & 0.5104 & 0.6779 & 4.0000 & 4.8346s & 14692.86 & 48.59 \\
& Ours      & \textbf{0.5281} & \textbf{0.7422} & 4.0000 & 4.9076s & \textbf{13261.07} & \textbf{43.86} \\
\midrule
\multirow{2}{*}{Qwen3-8B}
& SimpleMem & 0.4105 & 0.5981 & 4.0000 & 7.2318s & 12715.18 & 46.72 \\
& Ours      & \textbf{0.4459} & \textbf{0.6740} & 4.0000 & \textbf{6.6148s} & \textbf{11399.62} & \textbf{41.91} \\
\bottomrule
\end{tabular}
}
\end{table*}

The benefit is even clearer when viewed as context-budget efficiency. Table~\ref{tab:budget_efficiency_main} compares our method with $K=10$ retrieved memories against SimpleMem with $K=20$. On GPT-4.1-mini, our method with half the retrieved memory budget achieves higher F1 (0.4948 vs.\ 0.4896) and higher GT recall (0.7422 vs.\ 0.7325), while using roughly half the estimated context tokens. Qwen3-8B shows the same pattern: our method at $K=10$ exceeds SimpleMem at $K=20$ in both F1 (0.4031 vs.\ 0.3986) and GT recall (0.6740 vs.\ 0.6662), again with about half the context tokens. These results show that contextual reinstatement improves the ordering and contextual validity of retrieved memories, rather than relying on larger prompts or additional LLM reasoning. In practical long-term agent systems, this means the agent can reach the same or better performance while passing fewer memories to the generator.

\subsection{Qualitative Case Studies}
\label{app:case_studies}

We provide additional qualitative examples with the actual top retrieved memories from SimpleMem and RaMem. These cases show that context collapse is not caused by completely irrelevant retrieval. The baseline often retrieves memories that share entities or topics with the query, but they fail the query's evidence conditions because they belong to the wrong session or situation.

\begin{tcolorbox}[
    title={Case A: New Hobby on July 9},
    colback=gray!3,
    colframe=gray!60,
    fonttitle=\bfseries,
    breakable
]
\textbf{Query:} What new hobby did James become interested in on 9 July, 2022? \\
\textbf{Reference:} Extreme sports. \\
\textbf{SimpleMem answer:} surfing \quad
\textbf{RaMem answer:} extreme sports.

\vspace{0.4em}
\textbf{SimpleMem Top-3:}
\begin{enumerate}[leftmargin=1.5em,itemsep=0.15em]
    \item \textcolor{red}{Invalid}:
    \texttt{2022-03-20--2022-03-27}, gaming platform participation. James created a game avatar and joined a new gaming platform.
    \item \textcolor{red}{Invalid}:
    \texttt{2022-09-20--2022-10-03}, game streaming feedback. James streamed a game and received encouraging comments.
    \item \textcolor{red}{Invalid}:
    \texttt{2022-09-18--2022-09-20}, game streaming plans. James started streaming games.
\end{enumerate}
\textbf{GT rank:} 44.

\vspace{0.4em}
\textbf{RaMem Top-3:}
\begin{enumerate}[leftmargin=1.5em,itemsep=0.15em]
    \item \textcolor{green!50!black}{GT / valid}:
    \texttt{2022-07-09--2022-07-22}, extreme sports interest. James became interested in extreme sports and did rope jumping.
    \item \textcolor{red}{Invalid}:
    \texttt{2022-06-19--2022-07-09}, surfing experience. James went surfing.
    \item \textcolor{orange!80!black}{Related but topic-conflicting}:
    \texttt{2022-07-09--2022-07-22}, gaming tournament victory. James won an online gaming tournament.
\end{enumerate}
\textbf{GT rank:} 1.

\vspace{0.5em}
\textbf{Observation.}
SimpleMem retrieves memories that are related to James and activities, but all top-ranked evidence violates the query's recall conditions. RaMem reinstates the July 9 session context and places the verified memory at rank 1, turning retrieved memory from related fragments into contextually verifiable evidence.
\end{tcolorbox}

\begin{tcolorbox}[
    title={Case B: Vacation Episode vs. Later Partner-Related Memories},
    colback=gray!3,
    colframe=gray!60,
    fonttitle=\bfseries,
    breakable
]
\textbf{Query:} When did Evan get back from a vacation with his SO? \\
\textbf{Reference:} August 13, 2023. \\
\textbf{SimpleMem answer:} The context does not specify when Evan returned. \\
\textbf{RaMem answer:} August 13, 2023.

\vspace{0.4em}
\textbf{SimpleMem Top-3:}
\begin{enumerate}[leftmargin=1.5em,itemsep=0.15em]
    \item \textcolor{red}{Invalid}:
    \texttt{2023-12-31--2024-01-06}, accident and marriage news. Evan confirmed that he and his friends were fine after an accident.
    \item \textcolor{red}{Invalid}:
    \texttt{2024-01-06--2024-01-10}, family and partner support. Sam discussed Evan's partner and family support.
    \item \textcolor{red}{Invalid}:
    \texttt{2024-01-06--2024-01-10}, family support after marriage announcement. Evan mentioned support from extended family.
\end{enumerate}
\textbf{GT rank:} 18.

\vspace{0.4em}
\textbf{RaMem Top-3:}
\begin{enumerate}[leftmargin=1.5em,itemsep=0.15em]
    \item \textcolor{green!50!black}{GT / valid}:
    \texttt{2023-08-13--2023-08-15}, vacation details. Evan shared details about a recent vacation in Canada with his new partner.
    \item \textcolor{red}{Invalid}:
    \texttt{2023-12-31--2024-01-06}, accident and marriage news. Evan confirmed that he and his friends were fine after an accident.
    \item \textcolor{red}{Invalid}:
    \texttt{2023-11-21--2023-12-05}, mutual support. Evan thanked Sam for being there.
\end{enumerate}
\textbf{GT rank:} 1.

\vspace{0.4em}
\textbf{Observation.}
The baseline retrieves partner-related memories, but they belong to later sessions. RaMem identifies the vacation episode that satisfies the query's temporal and participant conditions.
\end{tcolorbox}

\begin{tcolorbox}[
    title={Case C: Gaming Topic vs. Correct Game Preference Episode},
    colback=gray!3,
    colframe=gray!60,
    fonttitle=\bfseries,
    breakable
]
\textbf{Query:} What game has Nate been playing nonstop with a futuristic setting and gameplay on October 9, 2022? \\
\textbf{Reference:} Cyberpunk 2077. \\
\textbf{SimpleMem answer:} Catan \quad
\textbf{RaMem answer:} Cyberpunk 2077.

\vspace{0.4em}
\textbf{SimpleMem Top-3:}
\begin{enumerate}[leftmargin=1.5em,itemsep=0.15em]
    \item \textcolor{red}{Invalid}:
    \texttt{2022-06-03--2022-06-05}, regional video game tournament. Nate shared a recent victory in a regional video game tournament.
    \item \textcolor{red}{Invalid}:
    \texttt{2022-08-22--2022-09-05}, gaming tournament and career. Nate won an international tournament and mentioned gaming as a career.
    \item \textcolor{red}{Invalid}:
    \texttt{2022-04-21--2022-05-02}, gaming tournament mention. Nate mentioned an upcoming gaming tournament.
\end{enumerate}
\textbf{GT rank:} 32.

\vspace{0.4em}
\textbf{RaMem Top-3:}
\begin{enumerate}[leftmargin=1.5em,itemsep=0.15em]
    \item \textcolor{orange!80!black}{Same session / related}:
    \texttt{2022-10-09--2022-10-21}, board game meeting. Nate met people playing the same board game.
    \item \textcolor{orange!80!black}{Same session / related}:
    \texttt{2022-10-09--2022-10-21}, game convention attendance. Nate attended a game convention and met new people.
    \item \textcolor{green!50!black}{GT / valid}:
    \texttt{2022-10-09--2022-10-21}, recent movie and game preferences. Nate mentioned playing \emph{Cyberpunk 2077}.
\end{enumerate}
\textbf{GT rank:} 3.

\vspace{0.4em}
\textbf{Observation.}
The baseline retrieves memories from the broad gaming topic, but they belong to earlier episodes. RaMem first narrows retrieval to the correct October 9 session, then includes the verified game-preference memory in the top retrieved evidence.
\end{tcolorbox}



\end{document}